%% file: main.tex
\documentclass[10pt,twocolumn,letterpaper]{article}

\usepackage[pagenumbers]{cvpr} 


\usepackage{fontawesome}
\usepackage{multirow}
\usepackage{arydshln}
\usepackage{comment}

\definecolor{cvprblue}{rgb}{0.21,0.49,0.74}
\usepackage[pagebackref,breaklinks,colorlinks,allcolors=cvprblue]{hyperref}

\def\paperID{*****} 
\def\confName{CVPR}
\def\confYear{2025}
\usepackage{xcolor}

\definecolor{red}{RGB}{255, 85, 85}  
\definecolor{blue}{RGB}{135, 206, 250} 
\definecolor{green}{RGB}{152, 251, 152}

\definecolor{lightgray}{gray}{0.9}
\newcommand{\CLA}[1]{{\color[HTML]{4472c4} \textbf{#1}}}
\newcommand{\CLB}[1]{{\color[HTML]{E76254} \textbf{#1}}}

\newcommand{\GroupA}[1]{{\color[HTML]{92ca2c} \textbf{#1}}}
\newcommand{\GroupB}[1]{{\color[HTML]{3c7daa} \textbf{#1}}}
\newcommand{\GroupC}[1]{{\color[HTML]{e64f04} \textbf{#1}}}
\newcommand{\GroupD}[1]{{\color[HTML]{f2c400} \textbf{#1}}}
\newcommand{\GroupE}[1]{{\color[HTML]{a757a7} \textbf{#1}}}

\title{Image Quality Assessment: From Human to Machine Preference}

\author{Chunyi Li$^{1,2,4}$$^{*}$, Yuan Tian$^{1,2}$$^{*}$, Xiaoyue Ling$^{1}$$^{*}$, Zicheng Zhang$^{1,2}$, Haodong Duan$^{2}$, Haoning Wu$^{3,4}$,\\ Ziheng Jia$^{1}$, Xiaohong Liu$^{1}$, Xiongkuo Min$^{1}$, Guo Lu$^{1}$$^{\dag}$, Weisi Lin$^{4}$, Guangtao Zhai$^{1,2}$$^{\dag}$\\
Shanghai Jiao Tong University$^{1}$, Shanghai AI Lab$^{2}$, Moonshot AI$^{3}$, Nanyang Technological University$^{4}$\\
}

\begin{document}

\twocolumn[{%
\renewcommand\twocolumn[1][]{#1}%
\maketitle
\begin{center}
    \centering
    \vspace{-2.4em}
    \includegraphics[width=1\linewidth]{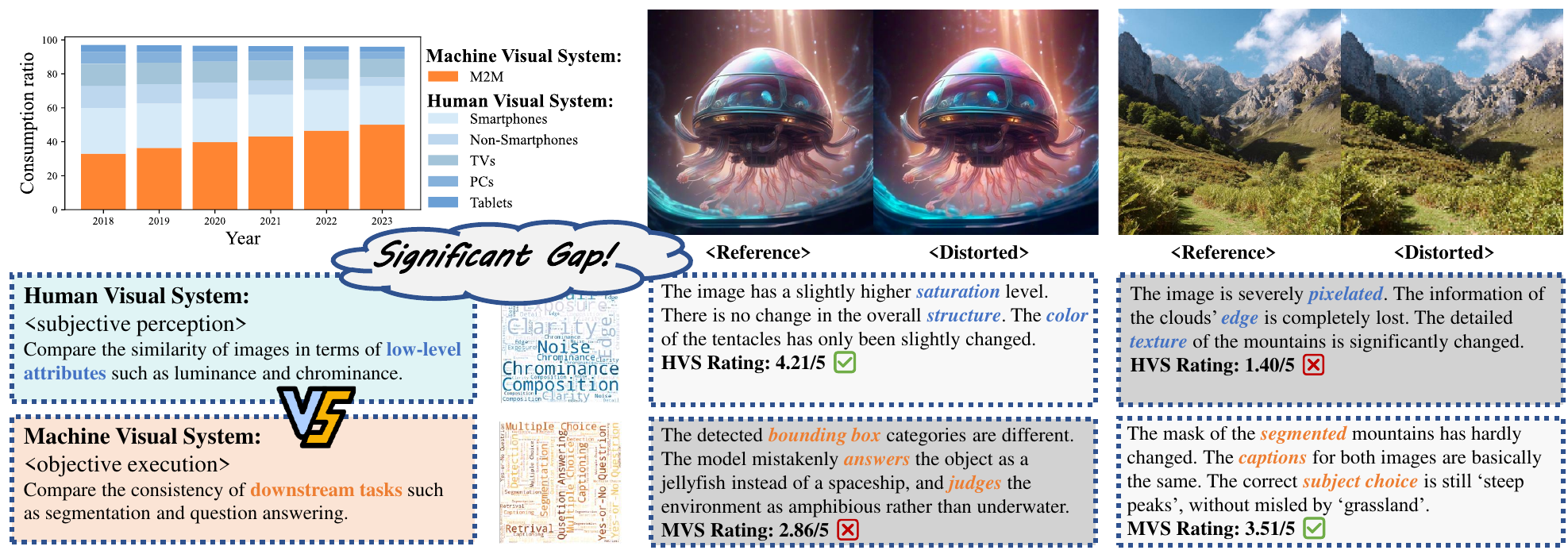}
    \vspace{-20pt}
    \captionof{figure}{The significant gap between the well-explored \CLA{Human Vision System (HVS)} and the emerging \CLB{Machine Vision System (MVS)}. Their different perception mechanisms leading to images that are subjectively satisfactory to humans not being applicable to the machines downstream tasks such as detection and question answering, vice versa.} 
    \label{fig:spotlight}
    
\end{center}%
}]

\newcommand\blfootnote[1]{%
\begingroup
\renewcommand\thefootnote{}\footnote{#1}%
\addtocounter{footnote}{-1}%
\endgroup
}

\begin{abstract}

Image Quality Assessment (IQA)\blfootnote{\vspace{-10pt}$^{*}$ Equal contribution. \quad $^{\dag}$ Corresponding authors.} based on human subjective preferences has undergone extensive research in the past decades. However, with the development of communication protocols, the visual data consumption volume of machines has gradually surpassed that of humans. For machines, the preference depends on downstream tasks such as segmentation and detection, rather than visual appeal. Considering the huge gap between human and machine visual systems, this paper proposes the topic: \CLB{Image Quality Assessment for Machine Vision} for the first time. Specifically, we (1) defined the subjective preferences of machines, including downstream tasks, test models, and evaluation metrics; (2) established the Machine Preference Database (MPD), which contains 2.25M fine-grained annotations and 30k reference/distorted image pair instances; (3) verified the performance of mainstream IQA algorithms on MPD. Experiments show that current IQA metrics are human-centric and cannot accurately characterize machine preferences. We sincerely hope that MPD can promote the evolution of IQA from human to machine preferences. Project page is on: \href{https://github.com/lcysyzxdxc/MPD}{https://github.com/lcysyzxdxc/MPD}.



\end{abstract}

\section{Introduction}
\label{sec:intro}

The rapid ascend of smart cities \cite{review:smart} and the flourishment of the Internet of Things (IoT) \cite{review:iot} over the past decade have fundamentally transformed the framework of the application end. According to the Cisco \cite{review:cisco} white paper, the number of Machine-to-Machine (M2M) connections first exceeded that of Machine-to-Human (M2H) in 2023, reaching 147 billion. Machines have gradually replaced humans and become the primary consumers of image/video data. In the field of image processing, the primary goal is to enhance the quality of processed images, ensuring they conform to human perception. Previously, the performance of a compression/reconstruction/generation algorithm \cite{my:misc,my:q-refine,my:g-refine} is often judged by its visual appeal to humans. This technology, known as Image Quality Assessment (IQA), comprehensively models the Human Visual System (HVS) to predict human subjective preferences. Consequently, with the evolution of application ends, IQA today has to transcend HVS and serve Machine Vision Systems (MVS) more effectively.

\begin{table*}[t]
\centering
    \renewcommand\arraystretch{1.25}
    \renewcommand\tabcolsep{4pt}
    \belowrulesep=0pt\aboverulesep=0pt
    \caption{Comparison of MPD with other IQA databases. As the first machine preference-oriented library, MPD has a larger image scale, a wider range of corruption scenarios, and more complete annotation content.}
    \label{tab:database}
    \vspace{-5pt}
    \resizebox{0.95\linewidth}{!}{
    \begin{tabular}{l|ccc|cc|ccr}
    \toprule
    {\multirow{2}{*}{Database}}          & \multicolumn{3}{c|}{Image} & \multicolumn{2}{c|}{Corruption} & \multicolumn{3}{c}{Annotation}                          \\ \cdashline{2-9}
              & Reference  & Distorted  & Resolution  & Types        & Strength        & Num   & Dimension & \multicolumn{1}{c}{Labeling Criteria}               \\ \midrule
    LIVE\cite{dataset:live}      & 29   & 779  & 768         & 5            & 5               & 25k   & 1   & Human preference on NSIs                  \\
    TID2013\cite{dataset:tid}       & 25   & 3k   & 512         & 24           & 5               & 514k  & 1   & Human preference on NSIs                  \\
    KADID-10K\cite{dataset:kadid10k} & 81   & 10k  & 1k          & 25           & 5               & 304k  & 1   & Human preference on NSIs                  \\
    CLIC2021\cite{dataset:clic}  & 585  & 3k   & 1k          & 10           & 3               & 484k  & 1   & Human preference on compressed images          \\
    NTIRE2022\cite{dataset:ntire2022} & 250  & 29k  & 288         & 40           & 5               & 1.13m & 1   & Human preference on super-resolution images      \\
    KonIQ-10K\cite{dataset:koniq} & -    & 10k  & 1k          & -            & -               & 1.21m & 5   & Human preference on NSIs                  \\
    AGIQA-3K\cite{dataset:agiqa-3k}  & -    & 3k   & 1k          & -            & -               & 125k  & 2   & Human preference on AIGIs                 \\
    NTIRE2024\cite{dataset:aigiqa20k} & -    & 20k  & 1k          & -            & -               & 420k  & 2   & Human preference on AIGIs                 \\
    \CLB{MPD} \textbf{(ours)}       & \textbf{1k}   & \textbf{30k}  & \textbf{1k}          & \textbf{30}           & \textbf{5}               & \textbf{2.25m} & \textbf{5}   & \CLB{Machine} \textbf{preference on} \CLB{NSIs, SCIs, and AIGIs} \\ \bottomrule
    \end{tabular}}
\end{table*}

However, there are significant disparities in the perceptual mechanisms between HVS and MVS. This discrepancy arises because HVS prioritizes the similarity between reference and distorted images, such as texture, structure, and color, whereas machines focus on the consistency of results in downstream tasks. As illustrated in Figure \ref{fig:spotlight}, for distortions resulting from factors like encoding/quantization, HVSs perceive a noticeable degradation in quality, yet the performance of tasks like segmentation/detection by machines remains unaffected. Conversely, some minor disturbances that the human eye barely notices can significantly disrupt the output of MVSs, making it difficult to migrate their evaluation algorithms.

What images do machines prefer? This remains an open question. For HVS, over the past two decades, IQA has witnessed significant development. Given the variations in image preferences among individuals, particularly across dimensions such as brightness and chroma, the scoring dimensions, the annotations from multiple subjects, and the integration of these annotations into global subjective preference labels have been well established. Hundreds of fine-grained databases can drive IQA algorithms that closely align with human perception. Conversely, MVS development is much slower, regardless of an end-to-end IQA algorithm, there is even no database that annotates images with an overall preference score of machines. This hinders the comprehensive evaluation of image processing metrics designed for machine vision, limiting the experience of application ends. Considering these issues, we first attempt to generalize the concept of IQA from HVS to MVS. Our contributions can be summarized as follows:

\begin{itemize} 
\item A innovative definition of the subjective preferences of machines. We propose a standardized process for collecting subjective labels from machines. Specifically, we identify the downstream tasks that determine the quality of machine perception, including four multimodal tasks related to Large Multimodal Models (LMMs) and three classic Computer Vision (CV) tasks. By referencing the multi-dimensional scoring mechanism of humans, we comprehensively evaluate the performance of machines in these tasks to obtain ground truth labels. 
\item A large-scale database named the Machine Preference Database (MPD). Using 15 general LMMs and 15 specialized CV models, we collected 2.25 million preference data points on 30,000 reference/distorted image pairs, establishing the first IQA database for machine vision. In particular, we conducted an in-depth analysis of each participating machine to explore the internal patterns of MVS perception and their differences with HVS. 
\item A complex benchmark test for existing IQA algorithms. Experiments on the MPD demonstrate that these algorithms designed for HVS cannot fully model MVS, whose predictions do not match machine ratings. After fine-tuning based on the MPD, their performance improves slightly, but still cannot characterize machine preferences, especially when facing mild distortion. 
\end{itemize}

\section{Related Works}

\begin{figure*}
\centering
\includegraphics[width = \linewidth]{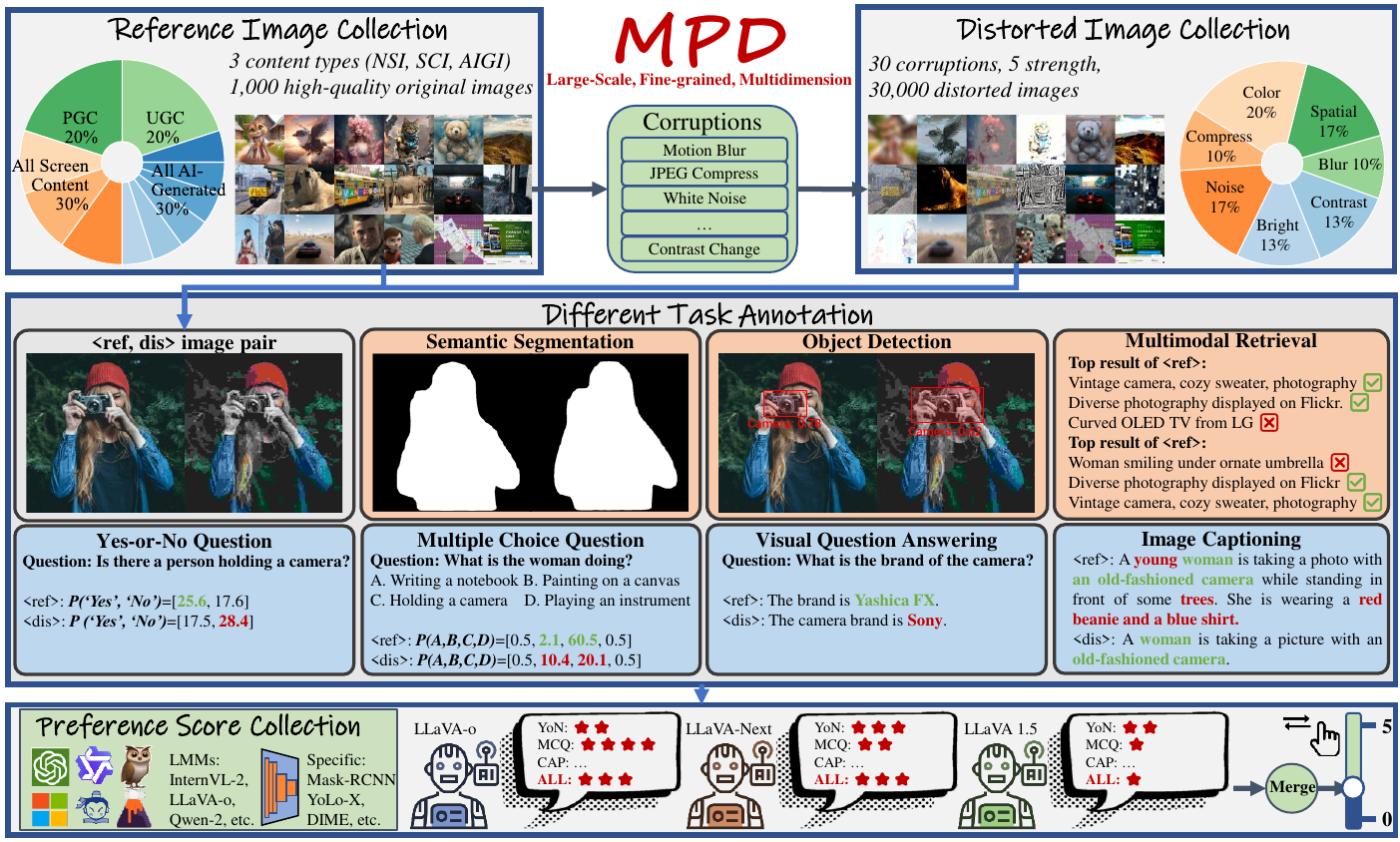}

\caption{Overview of Machine Preference Database (MPD), with 30k large-scale reference/distorted image pairs, meticulously annotated with 2.25M fine-grained preference score from the mean opinion of 15 mainstream LMMs and 15 specific CV models. It provides a multi-dimensional fidelity evaluation of various downstream machine vision tasks.}
\vspace{-3mm}
\label{fig:overview}
\end{figure*}

\subsection{Image Quality Assessment Database}

IQA datasets include Full Reference (FR) and No Reference (NR). FR datasets manually add corruption to reference images, whereas NR directly capture distorted images. Typically, these datasets focus on a single image type, such as Natural Scenes Images (NSIs) \cite{dataset:live,dataset:tid,dataset:kadid10k,dataset:clic,dataset:ntire2022}, Screen Content Images (SCIs) \cite{dataset:cct,dataset:scid,add:human}, and AI-Generated Images (AIGIs) \cite{dataset:agiqa1k,dataset:agiqa-3k,dataset:aigiqa20k,add:aigc}, with quality labels derived solely from human preferences as listed in Table \ref{tab:database}. On contrast, in the era where machines surpass humans as application ends, MPD stands out as the first machine preference-oriented library. It offers a larger image scale, a broader range of corruption scenarios, and a more comprehensive annotation context than other IQA databases.

\subsection{Machine Vision Downstream Task}


In the absence of a unified quality indicator, image processing algorithms, taking compression and restoration metrics for machine vision for example, are often in a fragmented state, with the performance of each algorithm being validated by only one or two downstream tasks. \cite{add:smr1,add:smr2} These tasks vary, as do the models used to perform them, making the evaluation criteria disparate and incomplete. For compression algorithms, validation indicators include the classification \cite{task:comp-class1,task:comp-class2}, image captioning \cite{task:comp-caption}, object detection \cite{task:comp-detection}, or semantic segmentation \cite{task:comp-segmentation} result before/after compression, with each indicator differing and lacking standardization or completeness. Restoration algorithms fare slightly better, with most consistently relying on object detection \cite{task:rest-1,task:rest-2,task:rest-3}. Nevertheless, some employ traditional Fast-RCNN \cite{det:faster-rcnn}, while others utilize more advanced models like ViTDet\cite{det:vitdet}. The variation in model bases further compromises the comprehensiveness and objectivity of the evaluation outcomes. Thus, there is an urgent need for a unified representation of machine preferences.

\section{Database Construction}
\subsection{Reference \& Distorted Image Collection}

Similar to the previous IQA database, we first selected 1,000 high-quality reference images and then artificially degraded their quality to obtain 30,000 pairs of reference and distorted images. To comprehensively represent the common content on the internet, we referred to the CMC-Bench \cite{dataset:cmc} and selected the three most common image categories: Natural Scene Images (NSIs), including User/Professional Generated Content (UGC/PGC) \cite{dataset:mscoco,dataset:clic}; Screen Content Images (SCIs), including webpages \cite{dataset:webpage}, games \cite{dataset:cgiqa}, and movies \cite{dataset:cct}; and AI-Generated Images (AIGIs), generated by six representative text-to-image models \cite{gen:dalle,gen:MJ,gen:Playground25,gen:pixart,gen:ssd-1b,gen:xl}. The collection process is shown in Figure \ref{fig:overview}. Both human viewers and IQA models have verified this image sequence to ensure that they are undistorted as reference images.

For the distorted images, we modeled the corruption caused by the current communication protocols, which were divided into 30 scenarios. Specifically, these corruptions were classified into seven categories: Blur, various types of unclear image; Luminance, global brightness changes; Chrominance, global color changes; Contrast, global changes in contrast/saturation; Noise, random noise of different distributions; Compression, codec algorithm like JPEG; and Spatial, local pixel-level changes.\footnote{For the definition of all these 30 corruptions, specific details, and examples, please refer to the appendix.} For each corruption, we defined five levels of intensity, from mild to severe. To fairly explore the impact of each corruption, we artificially control the distortion strength to ensure that quality degradation perceived by the human eye is aligned at the same level. Thus, for each reference image, we randomly selected the intensity to add all the corruptions mentioned above, resulting in 30,000 distorted images. The reference/distorted image pairs will then be annotated by machine subjects.

\subsection{Different Task Annotation}

In the past five years, the evaluation of image processing algorithms for machine vision has primarily focused on seven tasks: Regression, Classification, Visual Question Answering (VQA), Caption (CAP), Segmentation (SEG), Detection (DET), and Retrieval (RET). Among these, the first four tasks can be accomplished by general LMMs, whereas the latter three tasks necessitate specialized models due to their unique requirements. Note that the regression and classification tasks in this context are analogous to Yes-or-No (YoN) and Multiple-Choice Questions (MCQ) within LMMs. To this end, we enlisted five experienced human experts to design YoN/VQA/CAP questions and answers for 1,000 reference images. These questions are designed to be challenging, as illustrated in the example provided in Figure \ref{fig:overview}, especially in the MCQ sets where confusion options are introduced. This approach enables us to accurately measure the degree of quality degradation perceived by the machine.

We apply these seven tasks to both the reference and distorted images, measure the similarity of the output results, and derive scores for each of the seven sub-dimensions. Specifically, for \textbf{\textit{YoN}} tasks, we employ a logit-level evaluation approach based on softmax similarity, as previous LMM benchmarks \cite{bench:mcq-mmbench,bench:mcq-qbench} do. This entails LMMs confirming or denying the questions designed for each reference/distorted image pair. The confidence difference in the next token output, 'Yes' or 'No', reflects the difference in the perceptual quality degradation from the perspective of machines, namely a score $S_{YoN}$:
\begin{equation}
    S_{YoN}=|{\rm \sigma}(P_{dis}({\rm Yes,No}))-{\rm \sigma}(P_{ref}({\rm Yes,No}))|,
\end{equation}
where $P_{dis,ref}(\cdot)$ denotes the output probability of certain logits, normalized by the softmax function $\sigma(\cdot)$. \textbf{\textit{MCQ}} tasks follow a similar evaluation paradigm. We set a correct option and three confusing options (the options are usually no more than 4, see \cite{bench:mcq-vlmevalkit,bench:mcq-realworldqa,bench:mcq-mmbench}), and the model outputs the probability of each option as a four-element vector. The distance between the two vectors in space can reflect the perceptual quality score $S_{MCQ}$ for machines:
\begin{equation}
    S_{MCQ}={\rm cos}(P_{dis}({\rm A,B,C,D}),P_{ref}({\rm A,B,C,D})),
\end{equation}
where the distance is measured by cosine similarity. For \textbf{\textit{VQA}} tasks, we conduct sentence-level evaluation beyond the logits. LMMs will be required to answer a certain question in about 5 words, and the semantic difference between two output sentences will be the score $S_{VQA}$:
\begin{equation}
    S_{VQA}={\rm CLIP}(T_{dis},T_{ref}),
\end{equation}
where $T_{dis,ref}$ represents the whole output text, measured by the CLIP text encoder. \textbf{\textit{CAP}} tasks require a longer output, by describing the main subjects, actions, settings, and any notable objects or features of the image in about 40 words. Considering the remarkable text length, we use indicators dedicated to image captioning to analyze the differences between the two long sentences in $S_{CAP}$:
\begin{equation}
    S_{CAP}=\sum_{\rm eval} {\rm eval}(T_{dis},T_{ref}),
\end{equation}
where ${\rm eval} \in {\rm (BLEU,CIDEr,SPICE)}$ \cite{eval:bleu,eval:cider,eval:spice}, three most commonly-used captioning indicators.
\begin{figure}
\centering
\begin{minipage}[]{0.58\linewidth}
  \centering
  \centerline{\includegraphics[width = \textwidth]{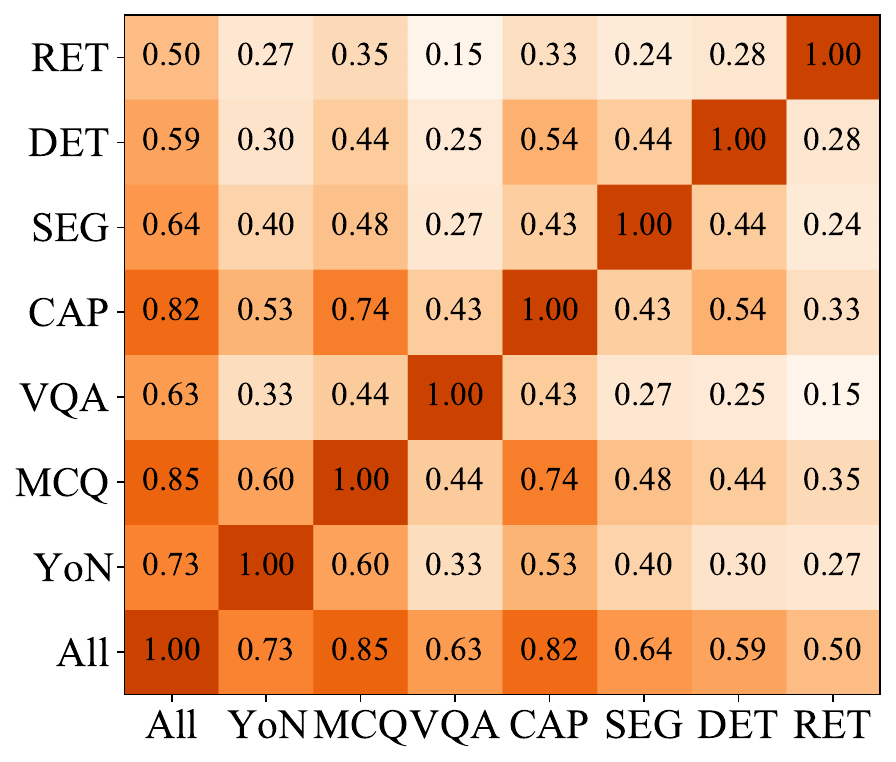}}
  \centerline{(a) Correlation matrix}\medskip
\end{minipage}
\vspace{-2mm}
\begin{minipage}[]{0.37\linewidth}
  \centering
  \centerline{\includegraphics[width = \textwidth]{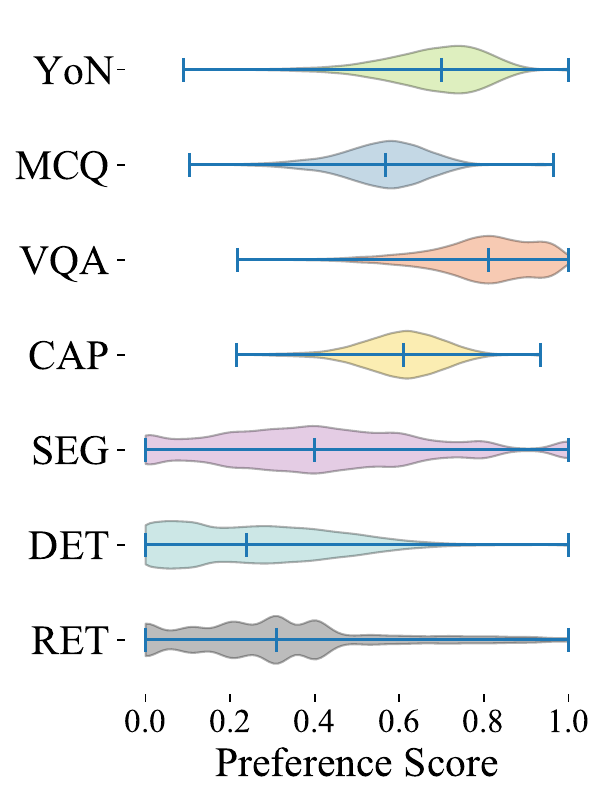}}
  \centerline{(b) Score distribution}\medskip
\end{minipage}
\vspace{-2mm}
\vspace{4mm}
\caption{Correlation between the general preference score and the score in seven different downstream tasks.}
\label{fig:task}
\end{figure}
Then, for \textbf{\textit{SEG}} tasks, referring to their common evaluation criteria \cite{review:seg}, we calculate the Intersection-over-Union (IoU) ratio in $S_{SEG}$:
\begin{equation}
    S_{SEG}={\rm IoU}(M_{dis},M_{ref}),
\end{equation}
where $M_{dis,ref}$ stands for binarized segmentation masks. \textbf{\textit{DET}} tasks will be more complex, whose results include both spatial information and category labels. Thus, we consider the detection failed when the categories are different, and only further calculate the IoU when they are consistent \cite{review:det}, as described in $S_{DET}$:
\begin{equation}
    S_{DET}={\rm Acc_1}(C_{dis},C_{ref})\cdot{\rm IoU}(B_{dis},B_{ref}),
\end{equation}
where $C_{dis,ref}$ denotes the category sequence ordered by confidence, and $B_{dis,ref}$ is the detected bounding box. Finally for \textbf{\textit{RET}} tasks, we merge three common Image-to-Text retrieval indicators \cite{review:ret}, namely Top-1/5/10 accuracy, to evaluate the retrieved sequence as $S_{RET}$:
\begin{equation}
    S_{RET}=\sum_{\rm i \in {1,5,10}} {\rm Acc_i}(C_{dis},C_{ref}).
\end{equation}
After the above process, we can obtain the preferences of LMMs in the first four dimensions, and the preferences of each specific CV model on its target task.
\begin{figure*}
\centering
\includegraphics[width=\linewidth]{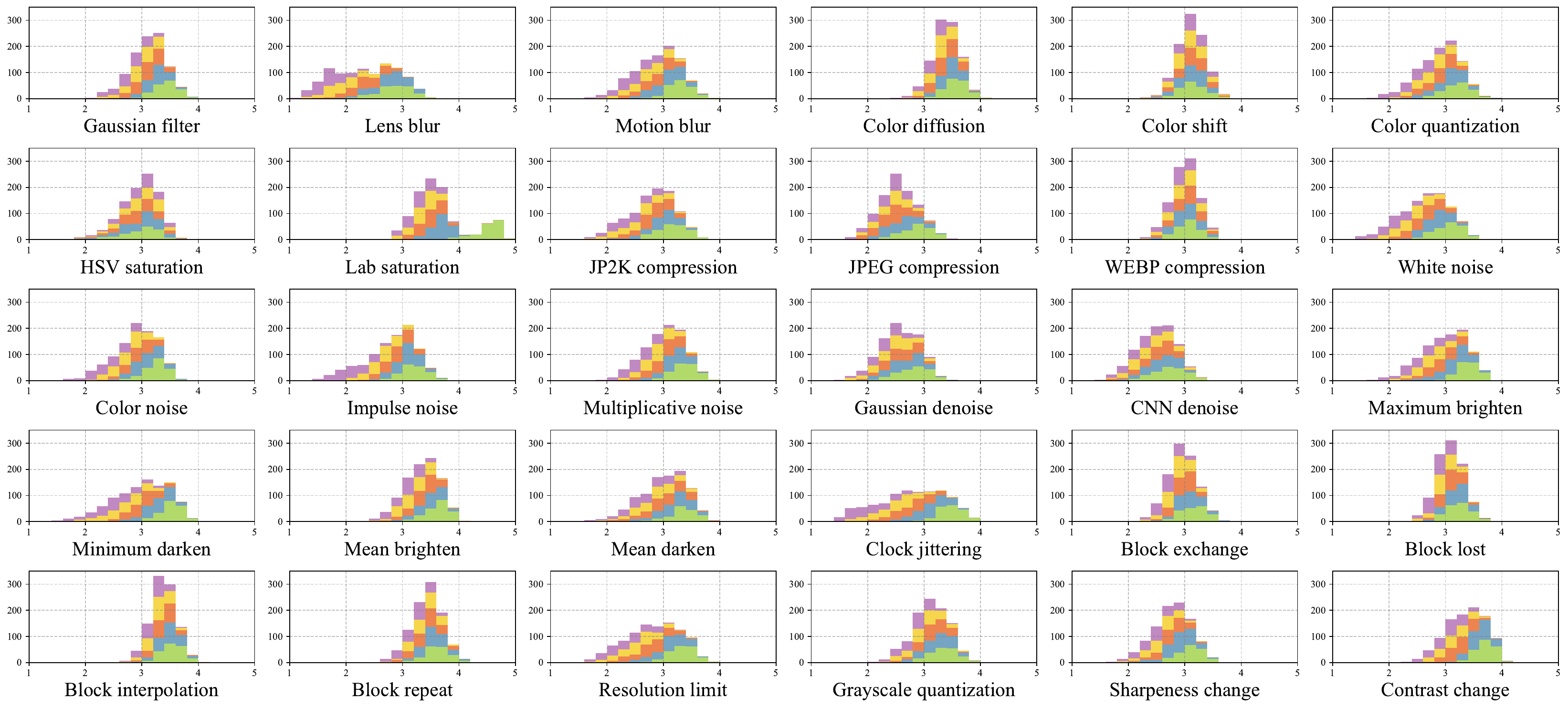}
\centering
\caption{MOS score of MPD, visualized in 30 corruption subsets. Different color denotes corruption strength \GroupA{Level 1}-\GroupB{Level 2}-\GroupC{Level 3}-\GroupD{Level 4}-\GroupE{Level 5}. Results show the sensitivity of the machines to each corruption varied significantly.}
\vspace{-4mm}
\label{fig:bar}
\end{figure*}

\subsection{Preference Score Collection}
Following the ITU human subjective annotation standards \cite{other:itu}, we adopt the same settings to obtain the machine Mean Opinion Score (MOS). Specifically, for YoN, MCQ, VQA, and CAP tasks, we use 15 general LMMs for inference, namely DeepseekVL \cite{model:deepseek}, InstructBLIP \cite{model:instructblip}, InternVL2 \citep{model:internvl2}, InternLM-XComposer2 \cite{model:internlmxcomposer2}, LLaVA1.5 \cite{model:llava15}, LLaVANext \cite{model:llavanext}, LLaVA-OneVision \cite{model:llavao}, Mantis \cite{model:mantis}, MiniInternVL \cite{model:miniintern}, MPlugOwl3 \cite{model:mplugowl3}, Ovis \cite{model:ovis}, Phi3.5 \cite{model:phi35}, Pangea \cite{model:pangea}, Qwen2-VL \cite{model:qwen2}, and Yi1.5 \cite{model:yi}. We set all temperature parameters as 0 to avoid unstable output to ensure all perceptual quality degradations come from distortion. Comprehensively considering popular trends and performance, we implement 5 models for each specific CV tasks, namely SEG: BoxInst \cite{seg:boxinst}, ConvNext \cite{seg:convnext}, Mask-RCNN \cite{seg:mask-rcnn}, SCNet \cite{seg:scnet}, and RTMDet (segmentation mode) \cite{det:seg:rtmdet}; DET: Dino\cite{det:dino}, Mask-RCNN \cite{det:faster-rcnn}, RTMDet (detection mode) \cite{det:seg:rtmdet}, ViTDet \cite{det:vitdet}, and Yolo-X \cite{det:yolox}; RET: DIME \cite{ret:dime}, InternVL (retrieval-based tuning) \cite{ret:internvl-tune}, LAVIS (retrieval-based tuning) \cite{ret:lavis-tune}, NAAF \cite{ret:naaf}, and VSE++ \cite{ret:vse++}. After obtaining scores for these 7 tasks, the evaluation can be divided into five dimensions, one dimension for each first four LMM tasks, and the last three tasks are unified into the last dimension, ensuring 15 subjects in each $(0,1)$ score for MOS. Thus merging all five dimensions into the $(0,5)$ MOS. Note that we normalize the score of each dimension to avoid where one dimension has a higher score while another is lower (e.g. VQA/CAP), which would lead to unfairness in the weight of the overall MOS.

In addition, considering the rapid development of LMM, we also apply 3 advanced LMMs, GPT4o \cite{model:gpt4}, GPT4o-mini \cite{model:gpt4}, InternLM-XComposer2d5 \cite{model:internlmxcomposer2}; and 4 early-stage LMMs, QWen1.5-VL \cite{model:qwen15}, VisualGLM \cite{model:visualglm}, Monkey \cite{model:monkey}, MPlugOwl2 \cite{model:mplugowl2} on certain tasks. We compared these results with the above-mentioned data in MPD, to prove the long-term usability of MPD through their consistency.
\begin{figure*}
\centering
\begin{minipage}[]{0.16\linewidth}
  \centering
  \centerline{\includegraphics[width = \textwidth]{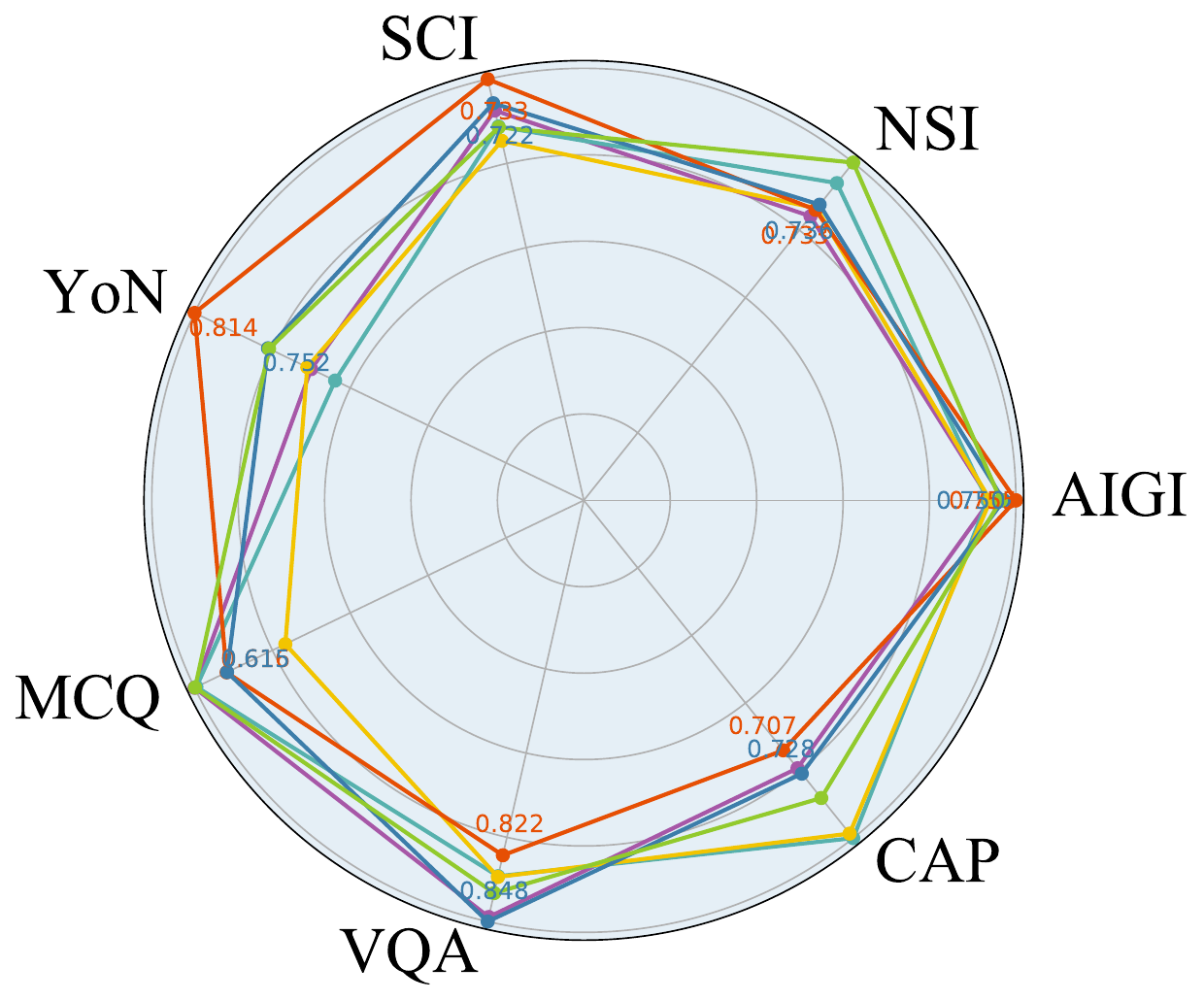}}
  \centerline{Strength 1}\medskip
\end{minipage}
\vspace{-2mm}
\begin{minipage}[]{0.16\linewidth}
  \centering
  \centerline{\includegraphics[width = \textwidth]{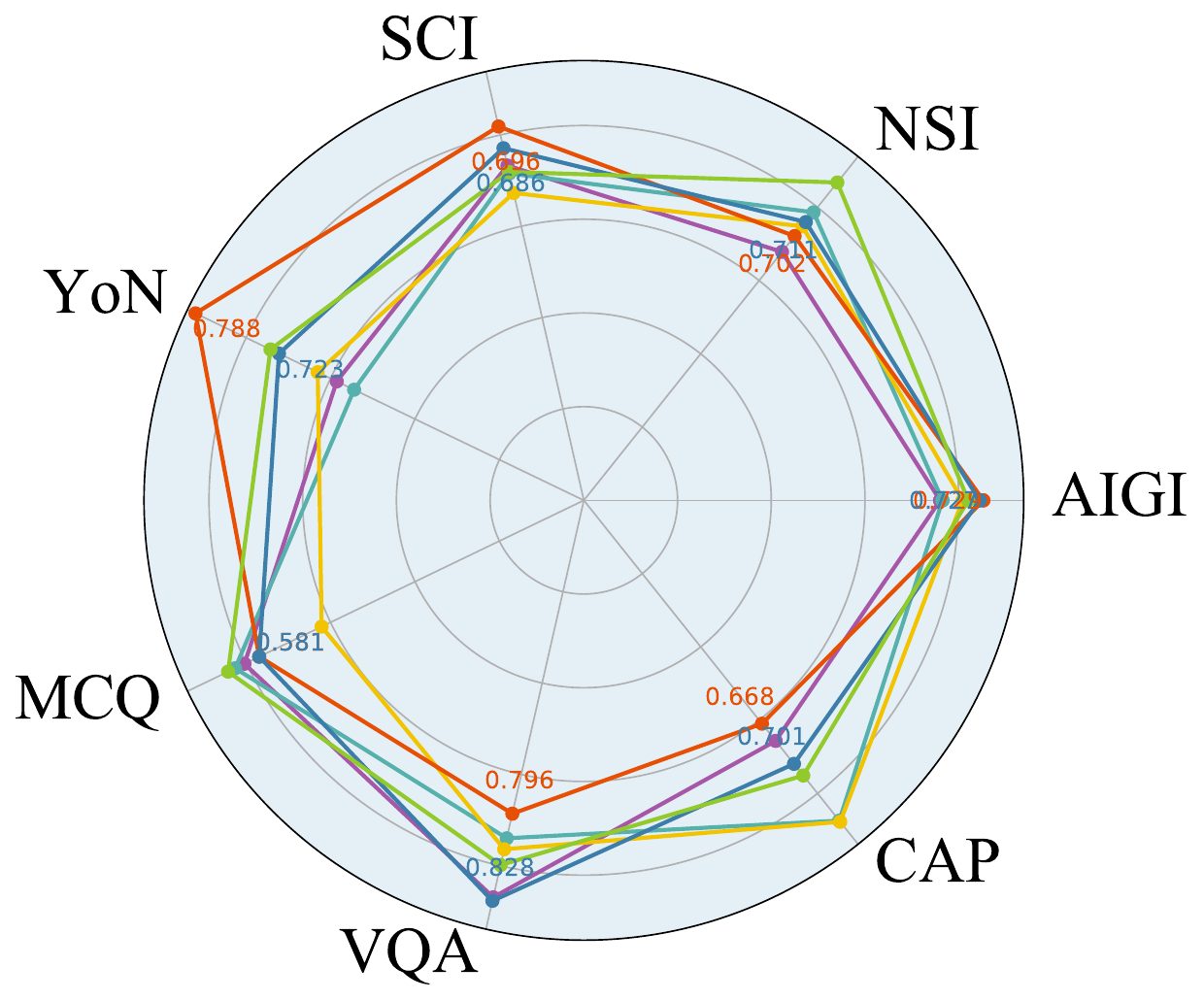}}
  \centerline{Strength 2}\medskip
\end{minipage}
\vspace{-2mm}
\begin{minipage}[]{0.16\linewidth}
  \centering
  \centerline{\includegraphics[width = \textwidth]{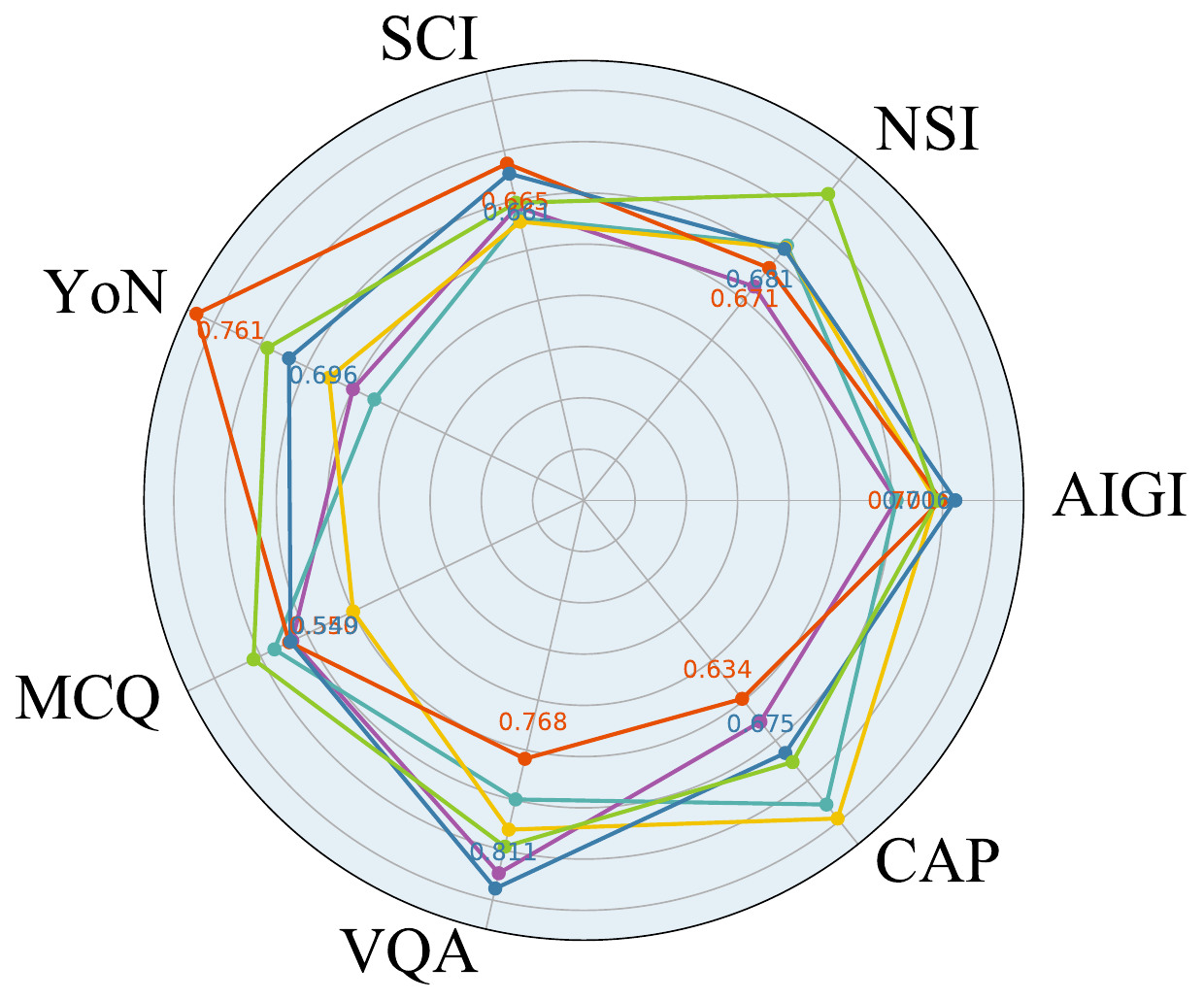}}
  \centerline{Strength 3}\medskip
\end{minipage}
\vspace{-2mm}
\begin{minipage}[]{0.16\linewidth}
  \centering
  \centerline{\includegraphics[width = \textwidth]{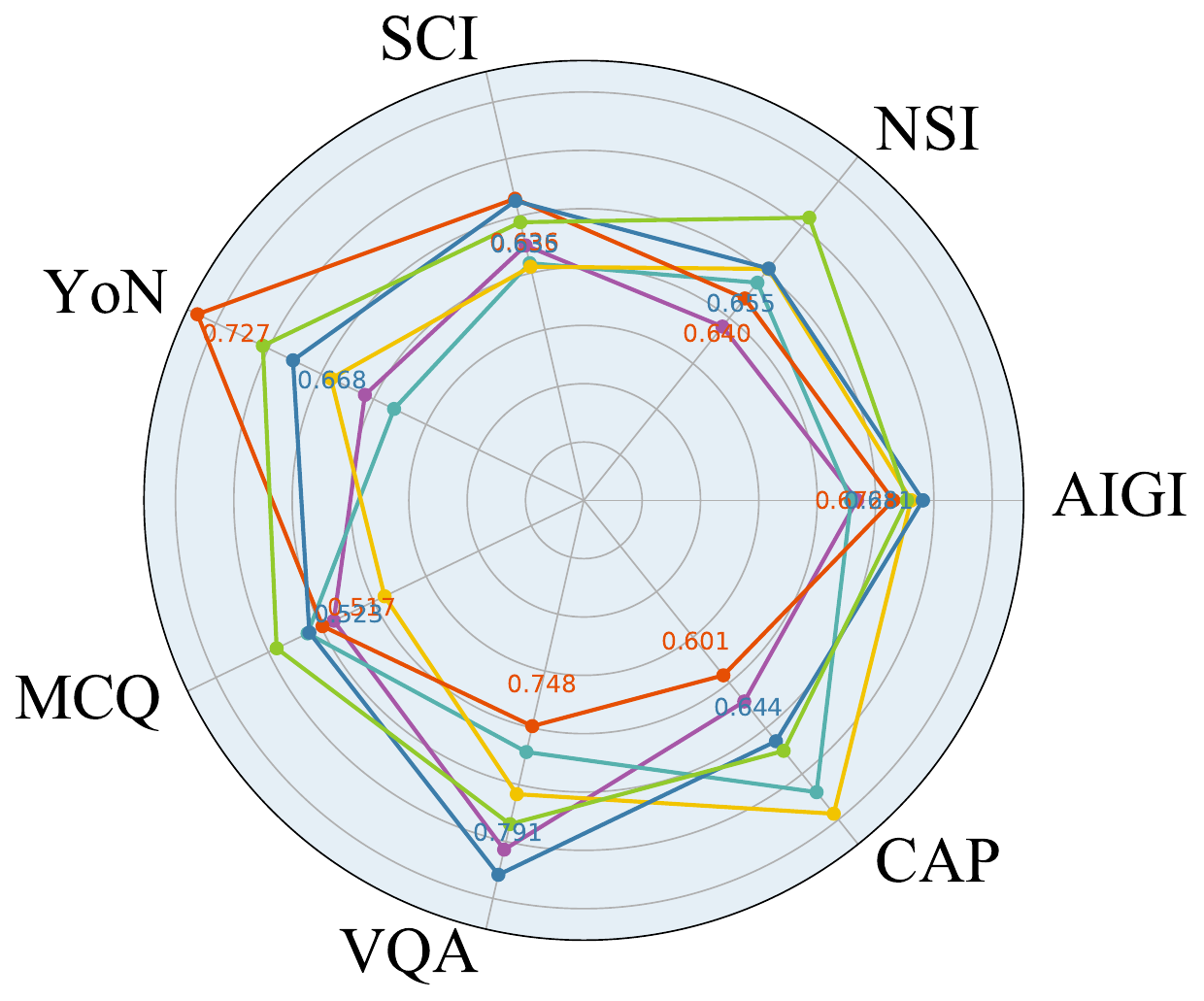}}
  \centerline{Strength 4}\medskip
\end{minipage}
\vspace{-2mm}
\begin{minipage}[]{0.16\linewidth}
  \centering
  \centerline{\includegraphics[width = \textwidth]{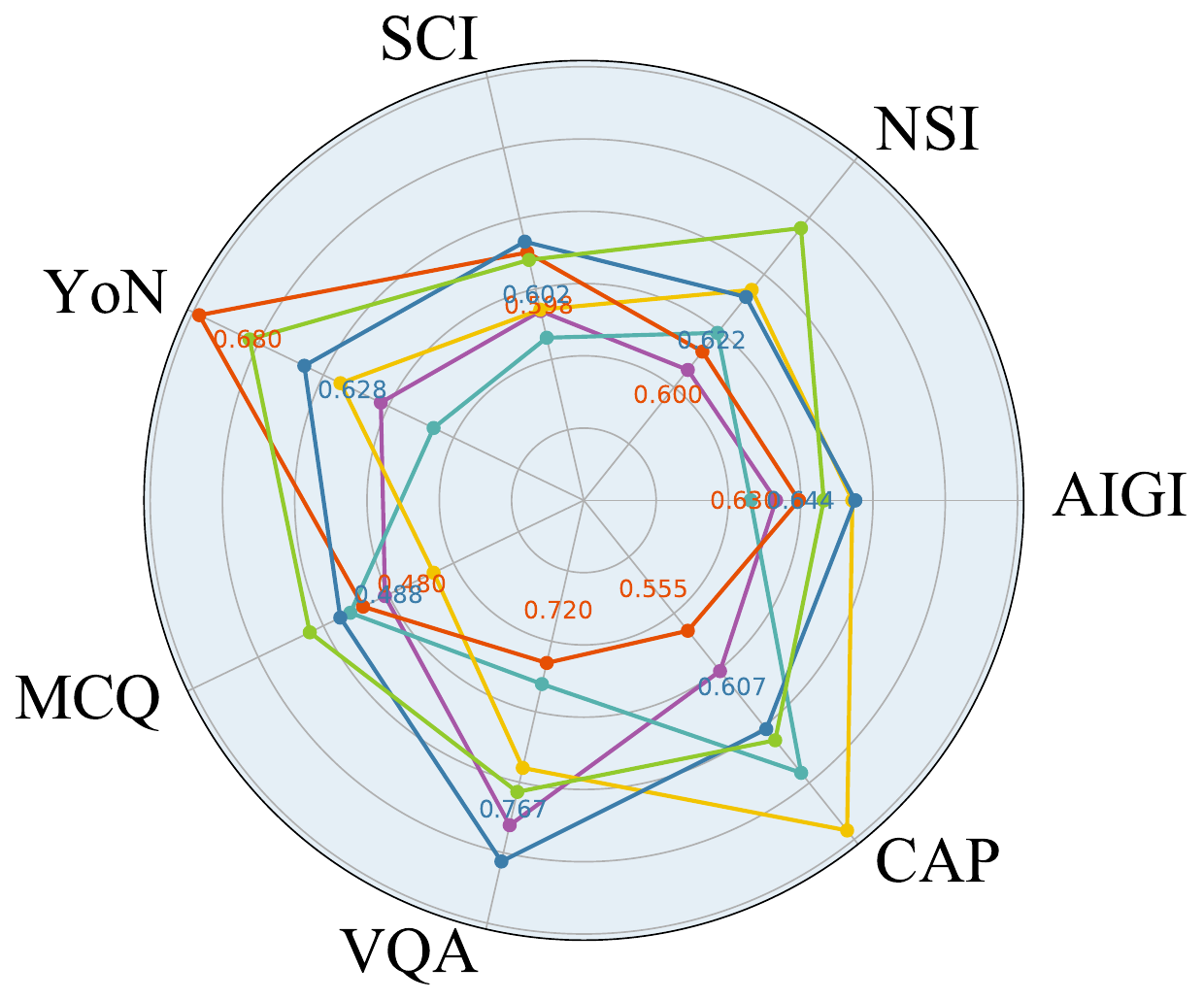}}
  \centerline{Strength 5}\medskip
\end{minipage}
\vspace{-2mm}
\begin{minipage}[]{0.11\linewidth}
  \centering
  \centerline{\includegraphics[width = \textwidth]{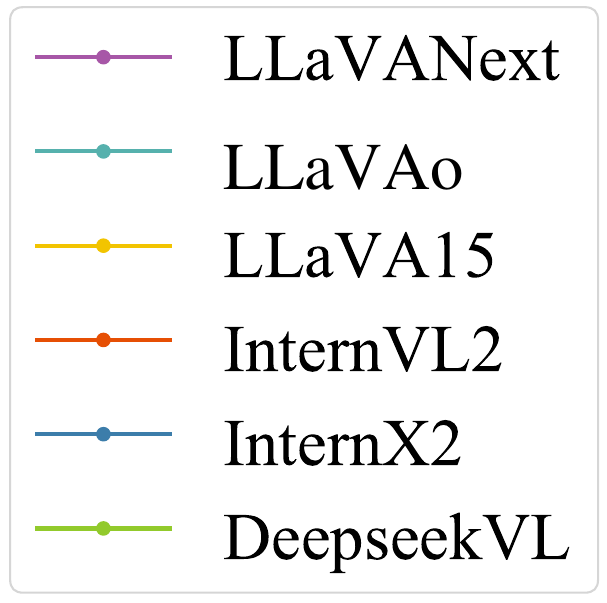}}
  \centerline{}\medskip
\end{minipage}
\vspace{-2mm}

\vspace{8mm}
\caption{Annotations from different LMMs subjects in MPD under 5 corruption strengths, including 3 content types and 4 tasks.}
\vspace{-2mm}
\label{fig:radar}
\end{figure*}
\begin{figure*}
\centering
\begin{minipage}[]{0.13\linewidth}
  \centering
  \centerline{\includegraphics[width = \textwidth]{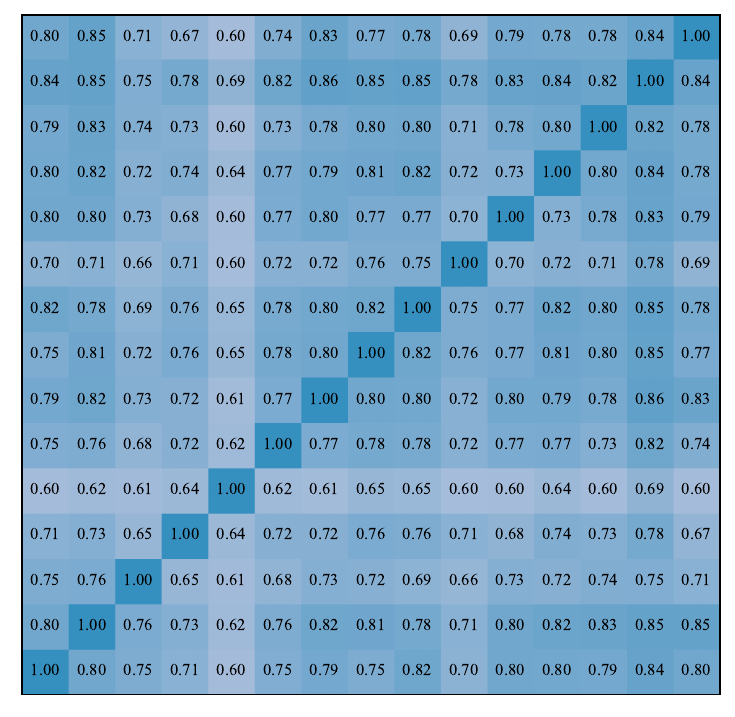}}
  \centerline{\CLA{Human} (0.76)}\medskip
\end{minipage}
\vspace{-2mm}
\begin{minipage}[]{0.13\linewidth}
  \centering
  \centerline{\includegraphics[width = \textwidth]{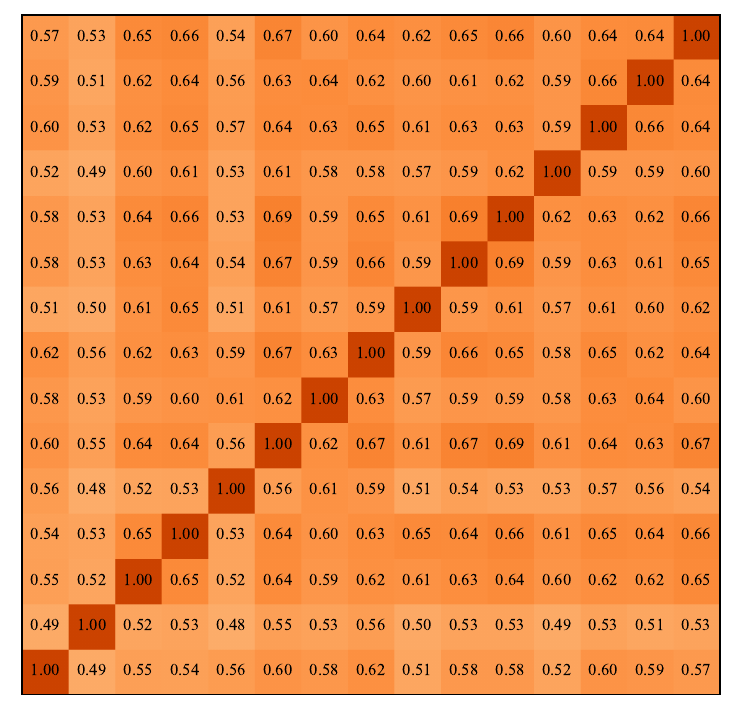}}
  \centerline{\CLB{MPD} (0.62)}\medskip
\end{minipage}
\vspace{-2mm}
\begin{minipage}[]{0.13\linewidth}
  \centering
  \centerline{\includegraphics[width = \textwidth]{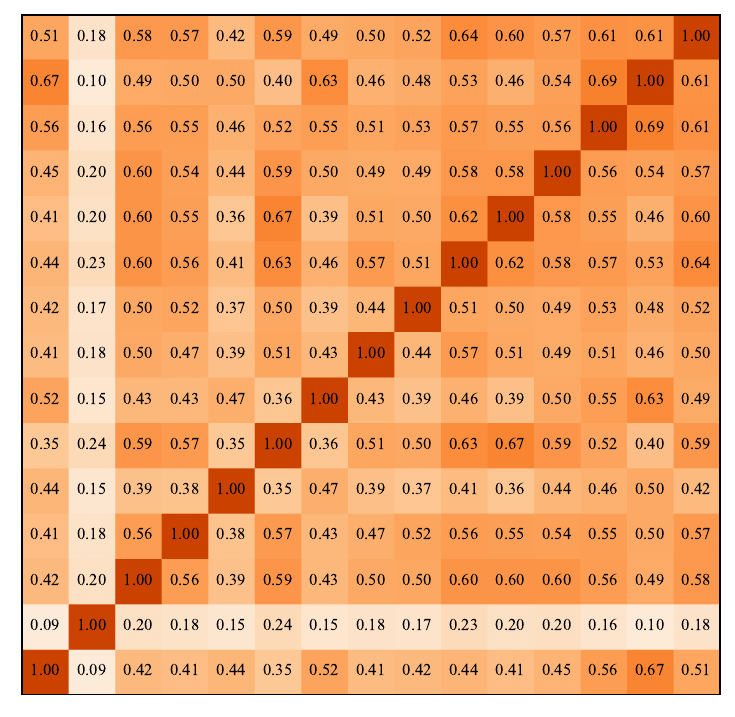}}
  \centerline{YoN (0.50)}\medskip
\end{minipage}
\vspace{-2mm}
\begin{minipage}[]{0.13\linewidth}
  \centering
  \centerline{\includegraphics[width = \textwidth]{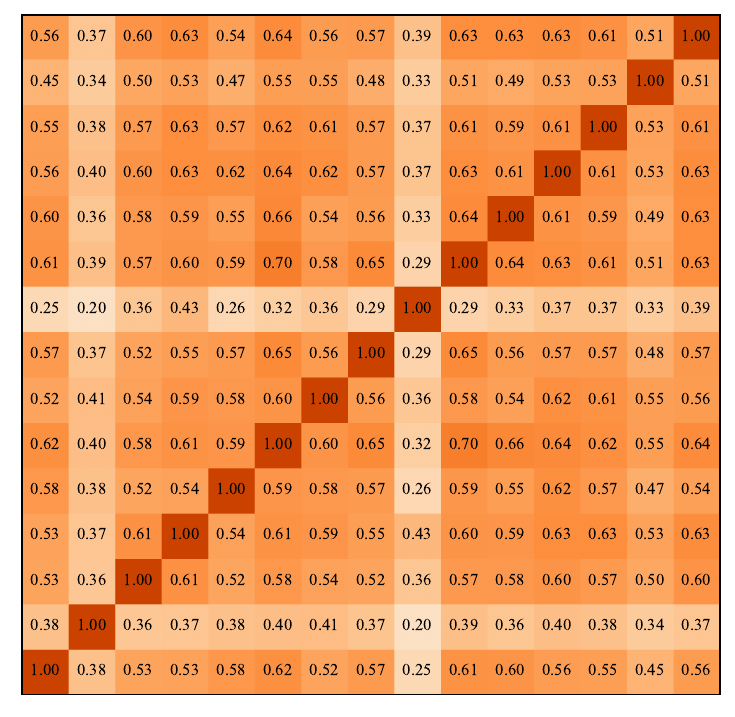}}
  \centerline{MCQ (0.55)}\medskip
\end{minipage}
\vspace{-2mm}
\begin{minipage}[]{0.13\linewidth}
  \centering
  \centerline{\includegraphics[width = \textwidth]{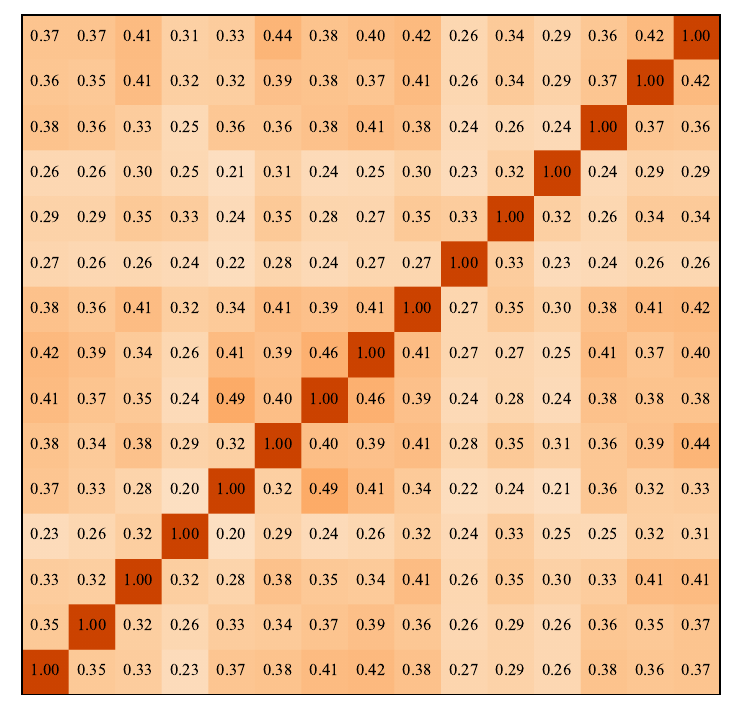}}
  \centerline{VQA (0.37)}\medskip
\end{minipage}
\vspace{-2mm}
\begin{minipage}[]{0.13\linewidth}
  \centering
  \centerline{\includegraphics[width = \textwidth]{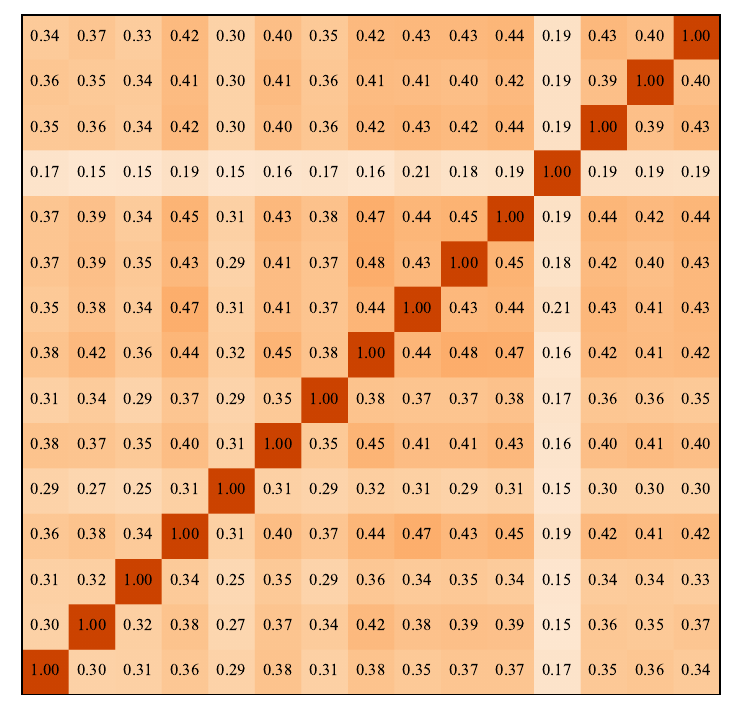}}
  \centerline{CAP (0.39)}\medskip
\end{minipage}
\vspace{-2mm}
\begin{minipage}[]{0.13\linewidth}
  \centering
  \centerline{\includegraphics[width = \textwidth]{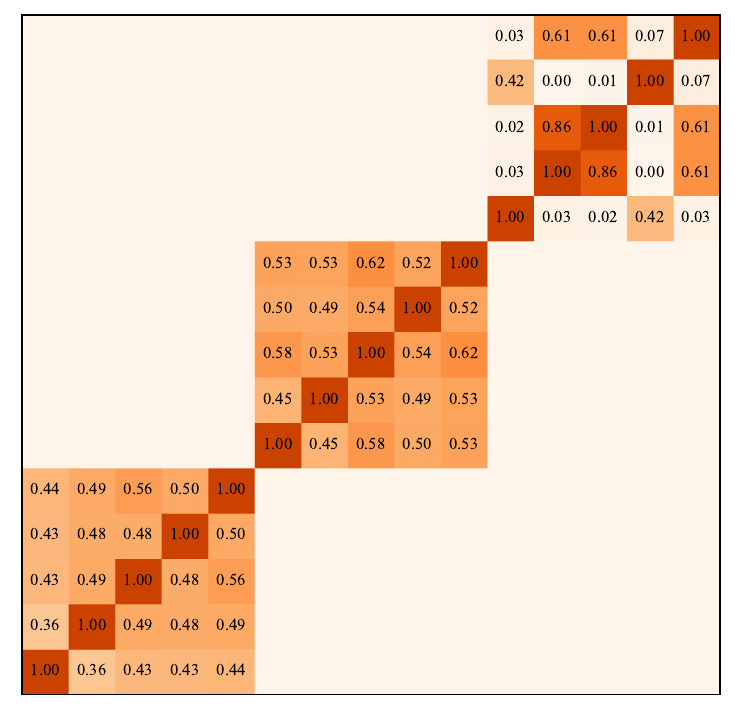}}
  \centerline{Other (0.53)}\medskip
\end{minipage}
\vspace{-2mm}

\vspace{10mm}
\caption{Correlation matrix of the proposed MPD in each dimension. A darker color denotes a higher SRCC, with mean SRCC attached below each matrix. The index of each matrix is arranged in the $\uparrow, \rightarrow$ directions, in the same order as Section 3.3. The correlation between \CLA{human} subjects is greater than \CLB{machines}, and the correlation of the overall MOS is greater than the score in each dimension.}
\label{fig:corr}
\end{figure*}
\section{Database Validation}

This section analyzes the preference data characteristics of MPD, to prove the usability of this database, including two perspectives: (1) the mean opinion of all machines; (2) the difference between each machine as a subject.

\textbf{Overall:} The MOS of MPD and the sub-scores of the seven tasks are illustrated in Figure \ref{fig:task}, with their correlation quantified by Spearman Rank-order Correlation Coefficient (SRCC). Notably, the inter-task correlation is relatively low, particularly for VQA and the three tasks that rely on specialized CV models. Given the current state of image processing algorithms for machine vision, each employing different tasks as performance metrics, the universality and robustness of such evaluation are inherently constrained. The data distribution further substantiates this argument, with YoN/MCQ/CAP being more evenly distributed, VQA exhibiting a higher concentration of high scores, and SEG/DET/RET clustering at lower scores. Moreover, the strong correlation between the total score and the other tasks is evident, with all SRCC surpassing 0.5, underscoring the reliability of the MPD, for various downstream tasks. The MOS distribution across 30 types of corruption and 5 strength levels is depicted in Figure \ref{fig:bar}. Results show for corruption scenarios that appear similar to HVS, the MVS perception mechanism diverges significantly. For instance, the machine is highly sensitive to `Lens blur' but rarely affected by `Mean brighten'. A typical example is the four Macro Block-related corruptions, which are nearly indistinguishable for HVS but exhibit significantly lower perception quality for MVS when encountering `Block exchange' compared to the other three. Furthermore, the impact of distortion intensity on the MVS perceptual quality varies markedly. For example, as the intensity level increases, the `Lab saturation' quality diminishes, while the `HSV saturation' quality remains unchanged. These findings underscore the differences between MVS and HVS, reinforcing the significance of establishing such a unified quality representation in MPD.

\textbf{Each machine:} We visualize the benchmark performance of six typical LMMs (as space limited) under different corruption strengths, as shown in Figure \ref{fig:radar}. For different content types, the scores of various machines on NSI/SCI/AIGI dimensions do not vary significantly. For different tasks, advanced models such as InternVL2 excel in dealing with distorted images in YoN, while traditional models like LLaVA1.5 can provide relatively stable results in CAP. As the corruption strength increases by one level, the scores of each dimension decrease by 0.01-0.02. Finally, for all machine subjects, we explored their internal differences in Figure 6. We manually conducted human annotation for 500 instances (the same 15 subjects) and compared the preference differences between humans and machines. The index of each matrix (i.e., the machine it represents) is arranged in the $\uparrow$ and $\rightarrow$ directions, in the same order as in Section 3.3. Overall, the SRCC between humans is 0.76, while that of machines is only 0.62, indicating that the divergence in machine image preferences is greater than that of humans. Further exploration of the preference scores for each dimension shows that the relatedness of machines in VQA and CAP tasks is the lowest. Additionally, there are a few subjects whose assessments differ significantly from other machines\footnote{The ITU human subjective annotation standard \cite{other:itu} for humans will exclude data with weak correlation with other subjects, but this is to prevent eye fatigue or malicious attacks during annotation. Given that the machine's inference process is completely objective, we will refer to the outputs of each machine rather than excluding them.}, such as InstructBlip in YoN, Mantis in MCQ, Pangea in CAP, and DIME/NAAF in RET. This is often due to special token outputs or multilingual mechanisms, which lead to insufficient robustness, resulting in a higher proportion of errors in distorted images compared to other models, thus leading to inconsistent results. In summary, the larger preference differences make IQA for MVS a more difficult problem than traditional IQA, but also a more valuable and promising research topic.

\begin{table*}[t]
\centering
    \caption{Using IQA metrics to predict machine preference, including baseline, FR, and NR algorithm, human visual system-based metrics are marked in {\faCheckCircle}. The first row represents two distortion intensities and three image content subsets, and the second row represents five scoring dimensions. [Keys: \CLB{Best}; \CLA{Second best}; \textbf{Baseline (BL)}; \colorbox{gray!20}{Inferior} to the baseline.]}
    \label{tab:iqa}
    \vspace{-8pt}
    \renewcommand\arraystretch{1.3}
    \belowrulesep=0pt\aboverulesep=0pt
    \resizebox{\linewidth}{!}{
\begin{tabular}{c|l|c|ccc|ccc|ccc|ccc|ccc}
    \toprule
                           & \multicolumn{1}{c|}{}                         & \multicolumn{1}{c|}{}                      & \multicolumn{3}{c|}{Severe Distortion}                                                                                    & \multicolumn{3}{c|}{Mild Distortion}                                                                                                              & \multicolumn{3}{c|}{NSI}                                                                                                  & \multicolumn{3}{c|}{SCI}                                                                                                  & \multicolumn{3}{c}{AIGI}                                                                                                 \\ \cdashline{4-18} 
    \multirow{-2}{*}{Type}     & \multicolumn{1}{c|}{\multirow{-2}{*}{Metric}} & \multicolumn{1}{c|}{\multirow{-2}{*}{HVS}} & SRCC$\uparrow$                                  & KRCC$\uparrow$                                  & PLCC$\uparrow$                                   & SRCC$\uparrow$                                  & KRCC$\uparrow$                                  & PLCC$\uparrow$                                                           & SRCC$\uparrow$                                  & KRCC$\uparrow$                                  & PLCC$\uparrow$                                   & SRCC$\uparrow$                                  & KRCC$\uparrow$                                  & PLCC$\uparrow$                                   & SRCC$\uparrow$                                  & KRCC$\uparrow$                                  & PLCC$\uparrow$                                   \\ \midrule
                               & PSNR                                         &                                           & 0.3873                                 & 0.2672                                 & 0.4664                                 & \textbf{0.3097}                        & \textbf{0.2127}                        & \textbf{0.6197}                                                & 0.4116                                 & 0.2820                                 & 0.4945                                 & 0.3135                                 & 0.2146                                 & 0.3329                                 & 0.5225                                 & 0.3684                                 & \textbf{0.5985}                        \\
    \multirow{-2}{*}{BL} & SSIM\cite{quality:ssim}                                         &                                           & \textbf{0.5969}                        & \textbf{0.4157}                        & \textbf{0.5537}                        & 0.2672                                 & 0.1828                                 & 0.2818                                                         & \textbf{0.6288}                        & \textbf{0.4466}                        & \textbf{0.5932}                        & \textbf{0.5415}                        & \textbf{0.3748}                        & \textbf{0.5393}                        & \textbf{0.5980}                        & \textbf{0.4278}                        & 0.5687                                 \\ \cdashline{1-18}
                               & AHIQ\cite{quality:ahiq}                                         &                                           & 0.8061                                 & 0.6153                                 & {\CLB{0.8183}} & {\CLA{0.5304}} & {\CLA{0.3685}} & \colorbox{gray!20}{\CLB{0.5111}} & 0.8447                                 & 0.6601                                 & {\CLA{0.8502}} & 0.7962                                 & 0.6040                                 & {\CLA{0.8470}} & 0.7996                                 & 0.6082                                 & \CLA{0.7966}                                 \\
                               & CKDN\cite{quality:ckdn}                                         & {\faCheckCircle}                                    & 0.7484                                 & 0.5560                                 & 0.7174                                 & \colorbox{gray!20}{0.0604}         & \colorbox{gray!20}{0.0405}         & \colorbox{gray!20}{0.1061}                                 & 0.8100                                 & 0.5988                                 & \colorbox{gray!20}{0.7670}         & 0.7203                                 & 0.5171                                 & 0.7661                                 & 0.6992                                 & 0.5102                                 & 0.6869                                 \\
                               & DISTS\cite{quality:dists}                                        &                                           & 0.7381             & 0.5443                                 & 0.7372                                 & 0.4616                                 & 0.3232                                 & \colorbox{gray!20}{0.4486}                                 & 0.7912                                 & 0.6060                                 & 0.8046                                 & 0.7275                                 & 0.5384                                 & 0.7429                                 & 0.7219                                 & 0.5365                                 & 0.7292                                 \\
                               & LPIPS\cite{quality:lpips}                                        & {\faCheckCircle}                                    & 0.6253                                 & \colorbox{gray!20}{0.4455}         & 0.5601                                 & \colorbox{gray!20}{0.0079}         & \colorbox{gray!20}{0.0092}         & \colorbox{gray!20}{0.0056}                                 & 0.6965                                 & 0.5132                                 & 0.6478                                 & 0.5618                                 & 0.3920                                 & 0.5686                                 & 0.6429                                 & 0.4518                                 & \colorbox{gray!20}{0.5481}         \\
                               & PieAPP\cite{quality:pieapp}                                       & {\faCheckCircle}                                    & \colorbox{gray!20}{0.4737}         & \colorbox{gray!20}{0.0163}         & \colorbox{gray!20}{0.2377}         & \colorbox{gray!20}{0.3030}         & \colorbox{gray!20}{0.2100}         & \colorbox{gray!20}{0.2349}                                 & 0.4536                                 & \colorbox{gray!20}{0.1187}         & \colorbox{gray!20}{0.2729}         & \colorbox{gray!20}{0.5173}         & \colorbox{gray!20}{0.0954}         & \colorbox{gray!20}{0.2135}         & \colorbox{gray!20}{0.4614}         & \colorbox{gray!20}{0.2009}         & \colorbox{gray!20}{0.3168}         \\
    \multirow{-6}{*}{FR}       & TOPIQ-FR\cite{quality:topiq}                                     & {\faCheckCircle}                                    & 0.7178                                 & 0.5252                                 & 0.7182                                 & 0.4248                                 & 0.2944                                 & \colorbox{gray!20}{\CLA{0.4841}}                                 & 0.7506                                 & 0.5614                                 & 0.7585                                 & 0.7243                                 & 0.5322                                 & 0.7186                                 & 0.6943                                 & 0.5063                                 & 0.7061                                 \\ \cdashline{1-18}
                               & ARNIQA\cite{quality:arniqa}                                       &                                           & {\CLB{0.8337}} & {\CLB{0.6426}} & {\CLA{0.8112}} & \colorbox{gray!20}{0.2388}         & \colorbox{gray!20}{0.1608}         & \colorbox{gray!20}{0.2754}                                 & {\CLA{0.8704}} & {\CLA{0.6690}} & 0.8478                                 & {\CLA{0.8220}} & {\CLA{0.6217}} & 0.8420                                 & {\CLB{0.8449}} & {\CLB{0.6578}} & 0.7954                                 \\
                               & CLIPIQA\cite{quality:CLIPIQA}                                      & {\faCheckCircle}                                    & 0.6498                                 & 0.4662                                 & 0.5901                                 & \colorbox{gray!20}{0.1822}         & \colorbox{gray!20}{0.1210}         & \colorbox{gray!20}{0.1797}                                 & 0.7670                                 & 0.5771                                 & 0.7141                                 & 0.6690                                 & 0.4751                                 & 0.6215                                 & 0.6877                                 & 0.4934                                 & \colorbox{gray!20}{0.5977}         \\
                               & DBCNN\cite{quality:DBCNN}                                        & {\faCheckCircle}                                    & \colorbox{gray!20}{0.5652}         & \colorbox{gray!20}{0.3908}         & 0.5696                                 & \colorbox{gray!20}{0.1334}         & \colorbox{gray!20}{0.0894}         & \colorbox{gray!20}{0.1594}                                 & 0.5978                                 & \colorbox{gray!20}{0.4181}         & 0.6219                                 & 0.5692                                 & 0.3974                                 & 0.6092                                 & \colorbox{gray!20}{0.5422}         & \colorbox{gray!20}{0.3810}         & \colorbox{gray!20}{0.5613}         \\
                               & HyperIQA\cite{quality:HyperIQA}                                     &                                           & {\CLA{0.8105}} & {\CLA{0.6189}} & 0.8012                                 & {\CLB{0.5388}} & {\CLB{0.3774}} & \colorbox{gray!20}{0.4712}                                 & {\CLB{0.8908}} & {\CLB{0.7074}} & {\CLB{0.8691}} & {\CLB{0.8273}} & {\CLB{0.6403}} & {\CLB{0.8601}} & {\CLA{0.8212}} & {\CLA{0.6242}} & {\CLB{0.8076}} \\
                               & NIMA\cite{quality:nima}                                         &                                           & 0.7562                                 & 0.5657                                 & 0.7858                                 & 0.4570                                 & 0.3171                                 & \colorbox{gray!20}{0.4342}                                 & 0.8153                                 & 0.6203                                 & 0.8224                                 & 0.7800                                 & 0.5881                                 & 0.8092                                 & 0.7846                                 & 0.5953                                 & 0.7622                                 \\
    \multirow{-6}{*}{NR}       & TOPIQ-NR\cite{quality:topiq}                                     & {\faCheckCircle}                                    & 0.6562                                 & 0.4616                                 & 0.6358                                 & \colorbox{gray!20}{0.1961}         & \colorbox{gray!20}{0.1318}         & \colorbox{gray!20}{0.1989}                                 & 0.7011                                 & 0.4823                                 & 0.6613                                 & 0.6645                                 & 0.4624                                 & 0.6353                                 & \colorbox{gray!20}{0.5461}         & \colorbox{gray!20}{0.3422}         & \colorbox{gray!20}{0.5249}   \\ \bottomrule     
    \end{tabular}}
    \resizebox{\linewidth}{!}{
\begin{tabular}{c|l|c|ccc|ccc|ccc|ccc|ccc}
    \toprule
                           & \multicolumn{1}{c|}{}                         & \multicolumn{1}{c|}{}                      & \multicolumn{3}{c|}{YoN}                                                                                    & \multicolumn{3}{c|}{MCQ}                                                                                                              & \multicolumn{3}{c|}{VQA}                                                                                                  & \multicolumn{3}{c|}{CAP}                                                                                                  & \multicolumn{3}{c}{Others}                                                                                                 \\ \cdashline{4-18} 
    \multirow{-2}{*}{Type}     & \multicolumn{1}{c|}{\multirow{-2}{*}{Metric}} & \multicolumn{1}{c|}{\multirow{-2}{*}{HVS}} & SRCC$\uparrow$                                  & KRCC$\uparrow$                                  & PLCC$\uparrow$                                   & SRCC$\uparrow$                                  & KRCC$\uparrow$                                  & PLCC$\uparrow$                                                           & SRCC$\uparrow$                                  & KRCC$\uparrow$                                  & PLCC$\uparrow$                                   & SRCC$\uparrow$                                  & KRCC$\uparrow$                                  & PLCC$\uparrow$                                   & SRCC$\uparrow$                                  & KRCC$\uparrow$                                  & PLCC$\uparrow$                                   \\ \midrule
                           & PSNR  & & 0.2753                                 & 0.1892                                 & 0.3467                                 & 0.3273                                 & 0.2239                                 & 0.4372                                 & 0.2067                                 & 0.1378                                 & 0.2120                                 & 0.4319                                 & 0.2993                                 & 0.4989                                 & 0.3798                                 & 0.2617                                 & 0.4683                                 \\
\multirow{-2}{*}{BL} & SSIM\cite{quality:ssim} &    & \textbf{0.3960}                        & \textbf{0.2728}                        & \textbf{0.3804}                        & \textbf{0.5089}                        & \textbf{0.3499}                        & \textbf{0.4880}                        & \textbf{0.2960}                        & \textbf{0.1988}                        & \textbf{0.2756}                        & \textbf{0.5550}                        & \textbf{0.3861}                        & \textbf{0.5447}                        & \textbf{0.5072}                        & \textbf{0.3491}                        & \textbf{0.4766}                        \\ \cdashline{1-18}
                           & AHIQ\cite{quality:ahiq}  &   & {\CLA{0.6017}} & {\CLA{0.4283}} & {\CLB{0.6212}} & {\CLA{0.7314}} & {\CLB{0.5374}} & {\CLB{0.7466}} & 0.4244                                 & 0.2924                                 & {\CLA{0.4583}} & {\CLA{0.7220}} & {\CLA{0.5341}} & {\CLA{0.7546}} & 0.6184                                 & 0.4396                                 & 0.6311                                 \\
                           & CKDN\cite{quality:ckdn}  & {\faCheckCircle}  & 0.5491                                 & 0.3799                                 & 0.5482                                 & 0.6866                                 & 0.4863                                 & 0.6596                                 & 0.4045                                 & 0.2763                                 & 0.3900                                 & 0.7044                                 & 0.5136                                 & 0.6991                                 & 0.5434                                 & 0.3822                                 & 0.5313                                 \\
                           & DISTS\cite{quality:dists} &   & 0.5267                                 & 0.3695                                 & 0.5394                                 & 0.6486                                 & 0.4625                                 & 0.6549                                 & 0.3512                                 & 0.2395                                 & 0.3676                                 & 0.6649                                 & 0.4833                                 & 0.6812                                 & 0.6203                                 & 0.4419                                 & {\CLB{0.6332}} \\
                           & LPIPS\cite{quality:lpips}  & {\faCheckCircle} & 0.4677                                 & 0.3252                                 & 0.4245                                 & 0.5517                                 & 0.3912                                 & 0.5169                                 & 0.3216                                 & 0.2169                                 & 0.2935                                 & 0.5620                                 & 0.3996                                 & 0.5473                                 & \colorbox{gray!20}{0.4927}         & \colorbox{gray!20}{0.3387}         & \colorbox{gray!20}{0.4242}         \\
                           & PieAPP\cite{quality:pieapp} & {\faCheckCircle} & \colorbox{gray!20}{0.3308}         & \colorbox{gray!20}{0.0173}         & \colorbox{gray!20}{0.1663}         & \colorbox{gray!20}{0.4310}         & \colorbox{gray!20}{0.1900}         & \colorbox{gray!20}{0.2087}         & \colorbox{gray!20}{0.2385}         & \colorbox{gray!20}{0.0132}         & \colorbox{gray!20}{0.1032}         & \colorbox{gray!20}{0.4220}         & \colorbox{gray!20}{0.0152}         & 0.2136                                 & \colorbox{gray!20}{0.3735}         & \colorbox{gray!20}{0.0209}         & \colorbox{gray!20}{0.1747}         \\
\multirow{-6}{*}{FR}       & TOPIQ-FR\cite{quality:topiq} & {\faCheckCircle} & 0.5011                                 & 0.3480                                 & 0.5118                                 & 0.6038                                 & 0.4232                                 & 0.6165                                 & 0.3700                                 & 0.2508                                 & 0.3899                                 & 0.6327                                 & 0.4545                                 & 0.6479                                 & 0.6199                                 & 0.4383                                 & 0.6281                                 \\ \cdashline{1-18}
                           & ARNIQA\cite{quality:arniqa}   & & {\CLB{0.6228}} & {\CLB{0.4394}} & {\CLA{0.6172}} & {\CLB{0.7342}} & {\CLA{0.5351}} & {\CLA{0.7242}} & {\CLA{0.4409}} & {\CLA{0.3026}} & 0.4454                                 & {\CLB{0.7656}} & {\CLB{0.5693}} & {\CLB{0.7664}} & {\CLB{0.6560}} & {\CLA{0.4566}} & 0.6309                                 \\
                           & CLIPIQA\cite{quality:CLIPIQA}  & {\faCheckCircle} & 0.4613                                 & 0.3167                                 & 0.4386                                 & 0.5090                                 & 0.3556                                 & 0.5117                                 & 0.2971                                 & 0.2028                                 & 0.3204                                 & 0.6379                                 & 0.4551                                 & 0.5673                                 & 0.5751                                 & 0.4048                                 & \colorbox{gray!20}{0.4747}         \\
                           & DBCNN\cite{quality:DBCNN}    & {\faCheckCircle} & \colorbox{gray!20}{0.3798}         & \colorbox{gray!20}{0.2567}         & 0.3969                                 & \colorbox{gray!20}{0.4793}         & \colorbox{gray!20}{0.3258}         & 0.4951                                 & \colorbox{gray!20}{0.2556}         & \colorbox{gray!20}{0.1719}         & 0.2783                                 & \colorbox{gray!20}{0.5240}         & \colorbox{gray!20}{0.3633}         & \colorbox{gray!20}{0.5388}         & 0.5085                                 & \colorbox{gray!20}{0.3465}         & 0.5093                                 \\
                           & HyperIQA\cite{quality:HyperIQA} & & 0.6010                                 & 0.4278                                 & 0.6066                                 & 0.7066                                 & 0.5155                                 & 0.7146                                 & 0.4322                                 & 0.2996                                 & 0.4419                                 & 0.7153                                 & 0.5234                                 & 0.7443                                 & {\CLA{0.6461}} & {\CLB{0.4598}} & {\CLA{0.6329}} \\
                           & NIMA\cite{quality:nima}     & & 0.5544                                 & 0.3904                                 & 0.5887                                 & 0.6893                                 & 0.5002                                 & 0.7036                                 & {\CLB{0.4460}} & {\CLB{0.3074}} & {\CLB{0.4775}} & 0.6313                                 & 0.4531                                 & 0.6842                                 & 0.5810                                 & 0.4106                                 & 0.6181                                 \\
\multirow{-6}{*}{NR}       & TOPIQ-NR\cite{quality:topiq} & {\faCheckCircle} & 0.4939                                 & 0.3358                                 & 0.4797                                 & 0.5469                                 & 0.3713                                 & 0.5370                                 & 0.3578                                 & 0.2437                                 & 0.3787                                 & \colorbox{gray!20}{0.5112}         & \colorbox{gray!20}{0.3375}         & \colorbox{gray!20}{0.5089}         & 0.5769                                 & 0.3934                                 & 0.5564                                   \\ \bottomrule     
    \end{tabular}}
    \vspace{-8pt}
\end{table*}

\section{Experiment}
\subsection{Experiment Setups}
We randomly partitioned the MPD into the train/val set, with 24,000 and 6,000 reference and distorted image pairs, respectively, according to an 8:2 ratio. For MVS, Figure \ref{fig:bar} illustrates significant variations in the perceptual quality of different corruptions. Given that machines exhibit a larger Just Noticeable Difference (JND) than humans, our corruption strength exceeds that of previous IQA databases for humans\footnote{See appendix for visualization in 5 corruption strength, where the Level 1 of our MPD can even ensemble Level 4 in KADID-10K\cite{dataset:kadid10k}.}. To conduct a realistic and reliable evaluation of machine preference, we designate the val set as `Severe Distortion' and select a subset of `Mild Distortion' to represent the corruption strength in the real world. The specific splitting mechanism is detailed in the Appendix. To benchmark the performance of quality metrics, three global indicators were employed: SRCC, Kendall Rank-order Correlation Coefficient (KRCC), and Pearson Linear Correlation Coefficient (PLCC), to evaluate the consistency between the objective quality score and the subjective MOS. Among these, SRCC and KRCC represent the prediction monotonicity, while PLCC measures the accuracy.

12 mainstream IQA metrics are implemented for comparison, which have all shown satisfying results in previous human-centric IQA tasks, including 6 FR metrics: AHIQ \cite{quality:ahiq}, CKDN \cite{quality:ckdn}, DISTS \cite{quality:dists}, LPIPS \cite{quality:lpips}, PieAPP \cite{quality:pieapp}, TOPIQ-FR \cite{quality:topiq}; and 6 NR metrics: ARNIQA \cite{quality:arniqa}, CLIPIQA \cite{quality:CLIPIQA}, DBCNN \cite{quality:DBCNN}, HyperIQA \cite{quality:HyperIQA}, NIMA \cite{quality:nima}, TOPIQ-NR \cite{quality:topiq}. Some methods specifically model the HVS, such as DBCNN, which has a dedicated stage for simulating human perception; others rely on the remarkable fitting ability of large-scale parameters, like the ViT in AHIQ. We have marked all HVS-based metrics as {\faCheckCircle}. Moreover, two naive quality indicators, PSNR and SSIM \cite{quality:ssim} are set as the baseline. We train the above 12 metrics on MPD with the learning rate as $10^{-5}$ for 50 epochs, under the default settings in pyiqa \cite{quality:topiq} toolbox and evaluate the performance on the two subsets above, three content types, and five scoring dimensions. We collect MPD preference annotations on 16$\times$NVIDIA A800 SXM4 80GB GPUs, and then conduct IQA training/validation on one GPU above.

\subsection{Experiment Result and Discussion}

Table \ref{tab:iqa} presents the performance of current advanced IQA models on the MPD. Overall, regardless of FR or NR, no metric is highly consistent with machine preference, although they are all good at predicting human preference. An interesting finding is that \CLB{the more accurately and completely a method models the HVS, the more difficult it is to evaluate the perceptual quality of MVS.} Considering that most of the current IQA works emphasize human vision, while machines have already surpassed humans to become the main consumer of visual signals, future IQA needs to find a way to balance these two goals.

For the two distortion subsets, fitting-based methods such as AHIQ/ARNIQA/HyperIQA can achieve SRCC/PLCC values exceeding 0.8 on Severe Distortion. However, this is not a challenging task. Under the same distortion conditions, they have even exceeded 0.95 on the HVS dataset \cite{quality:ahiq,dataset:tid}. The performance of Mild Distortion is even worse, and \CLB{no method can even outperform the simple PSNR, which proves that there is still a lot of room for optimization of IQA for machine preference.} For the three contents, after verifying them separately, the performance has slightly increased, which is caused by the internal quality differences of the contents. In general, predicting NSI quality is relatively simple, while SCI and AIGI are more difficult. This may be due to the fact that the initial weights of the IQA model have rich NSI priors. For the five dimensions, we can see that the results of MCQ and CAP are most consistent with machine preference; YoN and Others (three common CV tasks) are second; VQA is the worst. This is because VQA problems are usually changes in image details (such as the color of a button on a piece of clothing), which are more difficult for machines to detect than other tasks. In short, IQA for Machine preference still has great research potential, and it is necessary to develop more accurate and effective evaluation mechanisms in the future.

\begin{table}[t]
\centering
    \caption{Cross validation between human and machine preference database. For IQA metrics, \CLA{training on machine preference data} can reach a good result in evaluating human preference; but \CLB{training on human preference} is unacceptable for MPD. [Keys: \textbf{Best}; \colorbox{gray!20}{Lower} than 0.4.]}
    \label{tab:cross}
    \vspace{-8pt}
    \renewcommand\arraystretch{1.4}
    \belowrulesep=0pt\aboverulesep=0pt
    \resizebox{\linewidth}{!}{
\begin{tabular}{l|cccc|cccc}
\toprule
Val-set         & \multicolumn{2}{c}{\CLA{KADID-10K}} & \multicolumn{2}{c|}{\CLA{TID2013}} & \multicolumn{2}{c}{\CLB{MPD} (Mild)} & \multicolumn{2}{c}{\CLB{MPD} (Severe)} \\ \cdashline{1-9} 
Metric         & SRCC$\uparrow$                 & PLCC$\uparrow$                  & SRCC$\uparrow$                  & PLCC$\uparrow$                   & SRCC$\uparrow$                & PLCC$\uparrow$                 & SRCC$\uparrow$                 & PLCC$\uparrow$                  \\ \midrule
AHIQ\cite{quality:ahiq}     & 0.7950                & \textbf{0.7732}                & 0.7651                 & \textbf{0.7884}                 & \colorbox{gray!20}{0.1514}               & \colorbox{gray!20}{0.1846}               & 0.4753                & \colorbox{gray!20}{0.3518}                \\
CKDN\cite{quality:ckdn}     & \textbf{0.8599}                & 0.7282               & 0.6989                & 0.6029                 & \colorbox{gray!20}{0.1713}                 & \colorbox{gray!20}{0.2055}                 & \colorbox{gray!20}{0.3494}               & \colorbox{gray!20}{0.3228}               \\
DISTS\cite{quality:dists}    & 0.8239                & 0.8190                & 0.7231                 & 0.7759                 & 0.4105               & 0.4077               & 0.4616                & 0.4486                \\
LPIPS\cite{quality:lpips}    & 0.6184                & 0.5451                & 0.7054                & 0.6326                & \colorbox{gray!20}{0.2391}                 & \colorbox{gray!20}{0.3118}                 & 0.4683               & \colorbox{gray!20}{0.3711}               \\
PieAPP\cite{quality:pieapp}   & 0.4783                & \colorbox{gray!20}{0.3468}                & 0.6832                & 0.4543                & \colorbox{gray!20}{0.1285}               & \colorbox{gray!20}{0.0678}               & \colorbox{gray!20}{0.1791}                 & \colorbox{gray!20}{0.1235}                 \\
TOPIQ-FR\cite{quality:topiq} & 0.6768                & 0.6844                & \textbf{0.7782}                & 0.7473                & \textbf{0.5112}               & \textbf{0.5495}               & \textbf{0.6223}                 & \textbf{0.6685}                 \\
ARNIQA\cite{quality:arniqa}   & 0.7684                & 0.7023                & 0.7701                & 0.7205                & 0.4829                 & 0.4582                 & 0.5597               & 0.4849               \\
CLIPIQA\cite{quality:pieapp}  & 0.6286                & 0.5729                & 0.5237                & \colorbox{gray!20}{0.3910}                & \colorbox{gray!20}{0.1367}               & \colorbox{gray!20}{0.1478}               & \colorbox{gray!20}{0.3201}                & \colorbox{gray!20}{0.3203}                 \\
DBCNN\cite{quality:DBCNN}    & 0.5175                & 0.5351                & 0.4486                 & 0.5786                 & \colorbox{gray!20}{0.1359}               & \colorbox{gray!20}{0.1592}               & 0.4024                & \colorbox{gray!20}{0.3877}                \\
HyperIQA\cite{quality:HyperIQA} & 0.6650                & 0.6547                & 0.7024                 & 0.7313                 & \colorbox{gray!20}{0.0820}               & \colorbox{gray!20}{0.1074}               & 0.4519                & 0.4521                \\
NIMA\cite{quality:nima}     & 0.4795                & 0.4915                & 0.5694                 & 0.6423                 & \colorbox{gray!20}{0.1629}               & \colorbox{gray!20}{0.1775}               & 0.4709                & 0.4845                \\
TOPIQ-NR\cite{quality:topiq} & 0.5232                & 0.5077                & 0.5455                 & 0.6174                 & \colorbox{gray!20}{0.2388}               & \colorbox{gray!20}{0.2754}               & 0.4829                & 0.4582               \\ \bottomrule
\end{tabular}}
\vspace{-3mm}
\end{table}

\subsection{Cross Database Validation}

To further analyze the differences between MVS and HVS, we conducted cross-validation using MPD and two of the most commonly used human-oriented datasets, TID2013 \cite{dataset:tid} and KADID-10K \cite{dataset:kadid10k}, employing the same training mechanism as in the previous chapter. We observed that after training on MPD, most IQA metrics still possess a certain ability to predict human preferences, with SRCC values around 0.7 to 0.8; conversely, after training on the other two human-centric databases, IQA metrics struggle to predict machine preferences, with results on MPD not even reaching 0.4. \CLB{Machine preference data will not affect human tasks, but the opposite will.} This finding holds certain guiding significance for future IQA, particularly for a human-machine-friendly paradigm.

\begin{figure}
\centering
\begin{minipage}[]{0.48\linewidth}
  \centering
  \centerline{\includegraphics[width = \textwidth]{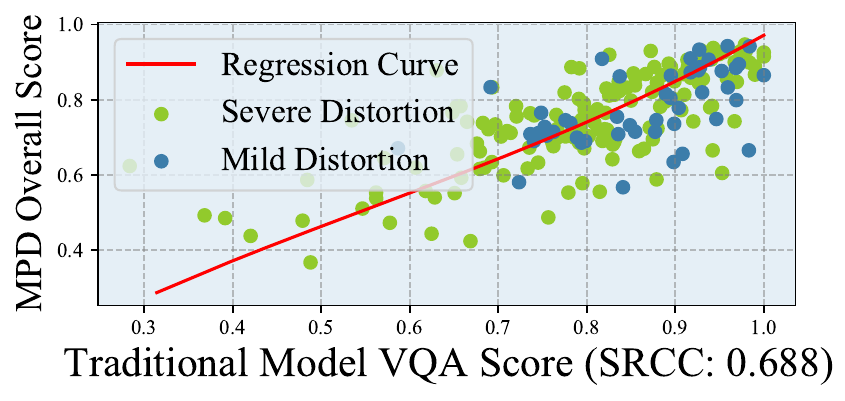}}
\end{minipage}
\begin{minipage}[]{0.48\linewidth}
  \centering
  \centerline{\includegraphics[width = \textwidth]{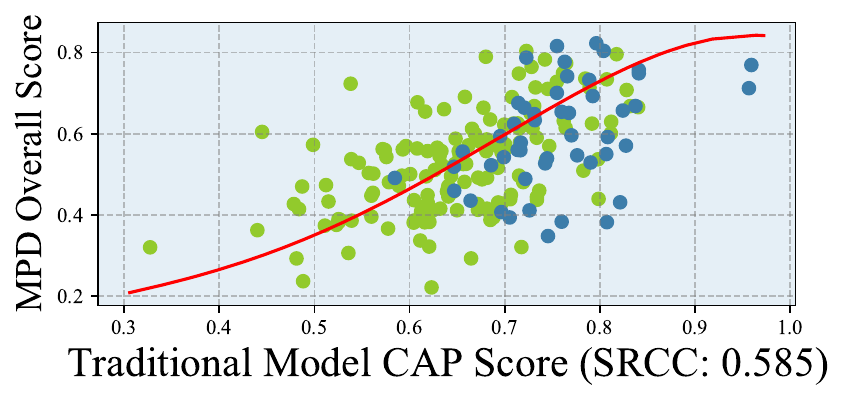}}
\end{minipage}
\vspace{-2mm}

\begin{minipage}[]{0.48\linewidth}
  \centering
  \centerline{\includegraphics[width = \textwidth]{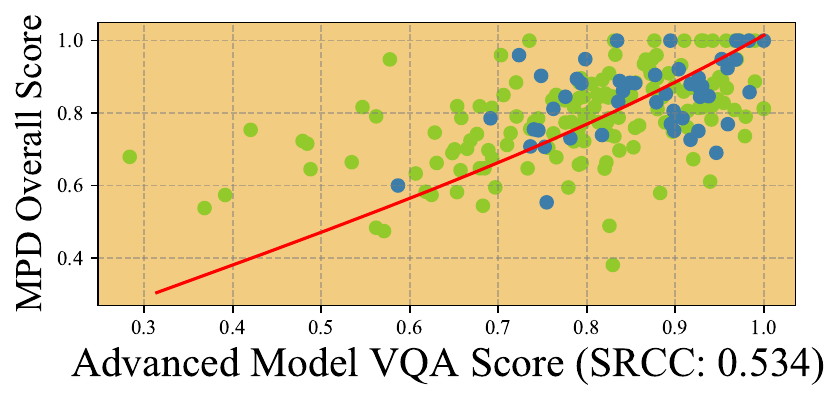}}
\end{minipage}
\begin{minipage}[]{0.48\linewidth}
  \centering
  \centerline{\includegraphics[width = \textwidth]{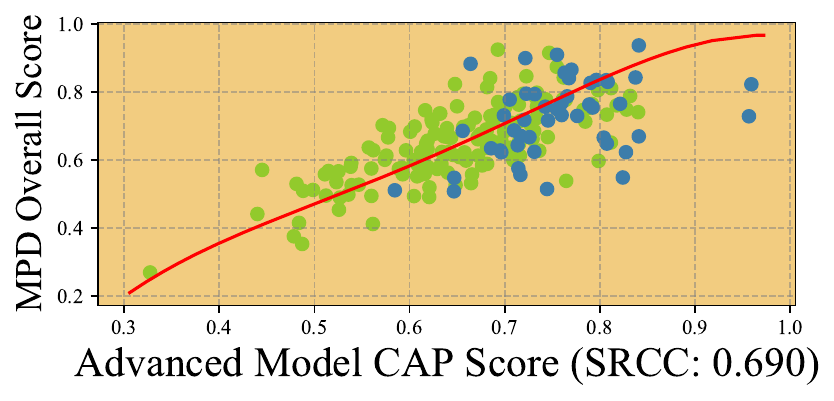}}
\end{minipage}
\vspace{-2mm}

\vspace{2mm}
\caption{The long-term usability of MPD. The preference scores of MPD are consistent with four early-stage LMMs and three advanced LMMs (marked as \CLA{Traditional}/\CLB{Advanced} Models in the background). Their high SRCC proves that MPD can still characterize the perceptual quality of LMMs in the near future.}
\label{fig:time}
\end{figure}

\subsection{Long-term Availability Validation}

The machine subjects in MPD are categorized into two groups: common CV models and LMMs. The former models, such as the separation/detection benchmark method, have been largely finalized and generally recognized. In contrast, LMMs are still under rapid evolvment. It is conceivable that MPD may not be able to capture the LMMs that will emerge in the future, or with the update of MPD, the preference of previous LMMs may be sacrificed. Thus, we include 3 advanced and 4 early-stage LMMs as Sec 3.3 listed, integrating them into two databases similar to MPD. Given the inconsistencies in LMM preference on VQA and CAP tasks, as illustrated in Figure \ref{fig:corr}, we have verified the consistency of MPD ground truth labels with advanced/early-stage labels. For each image, the relationship between these two labels is depicted in Figure \ref{fig:time}. The findings indicate that MPD exhibits a high correlation with early models on VQA and with advanced models on CAP. Generally, the SRCC is around 0.6, which is considered acceptable. Considering the GPT-4o surpasses open-source models, we anticipate that MPD is able to characterize LMMs preferences for images for at least a year. 
In the future, We will continue to update MPD by periodicity introducing the latest LMMs to avoid this limitation.

\section{Conslusion}

This paper extends the application of IQA from human to machine vision for the first time and establishes the first large-scale, fine-grained, multi-dimensional quality database for machines, namely MPD. We select 15 LMMs and 15 common CV models as subjects, show them 30k reference/distorted image pairs, and summarize their performance on 7 downstream tasks as quality scores. In-depth analysis of the MPD content and IQA benchmark experiments have proved the huge gap between machine and human visual systems, and current human-centric IQA methods cannot accurately predict machine preferences. As the visual data consumption volume of machines has already surpassed humans, we hope that MPD can inspire effective IQA metrics to serve the upcoming machine-centric era.


\clearpage
\maketitlesupplementary
\appendix

\section{Limitation and Broader Impact}

\textbf{[Limitation 1]}: The main limitation of this study lies in the rapid iteration of LMMs. The study of human-centric IQA is easier because the preferences of different humans do not vary greatly and do not change rapidly over time. The definition of `beauty' in human society typically iterates over centuries. However, mainstream LMMs may be updated annually, and their preferences will also change accordingly. Given that machines are already the primary consumers of visual data, this issue must be addressed. As mentioned in the main text, we can ensure that the annotated data of MPD and the LMMs within a year are relatively consistent. For future LMMs, we will continue to update the machine subjects to ensure their long-term usability of MPD.
\\
\textbf{[Limitation 2]}: The calculation method of MOS is also one of the limitations of this study. To comprehensively consider the preferences of machines, we consider seven tasks simultaneously and directly sum $YoN+MCQ+VQA+CAP+Others$ to derive MOS scores. However, in practical applications, their importance is not necessarily the same, and developers may only focus on certain tasks. In response, we will make the scores of each sub-dimension public, allowing developers to define weights freely. For example, an image compression method aimed at image captioning could increase its share in MOS by only focusing on the $CAP$ dimension during training. Thus, the broad applicability of MPD can be guaranteed. (However, the selection of weights still requires manual adjustment.)
\\
\textbf{[Broader Impact]}: For the field of IQA, this study has fundamentally overturned the human-centric paradigm that has dominated the past 20 years. MPD has for the first time clearly defined the tasks of `IQA for Machine Vision.' We believe it will inspire future machine-centric IQA methods. For the entire machine vision research community, this study has ended the situation where different models are being operated and evaluated independently. With the vigorous development of machine-oriented image processing algorithms (such as the rapid development of image/video coding for machines in the past five years), MPD unifies its evaluation criteria. Unlike the previous situation where one side used segmentation and the other used detection, MPD makes a fair, broad, and credible evaluation possible. Considering the important position of machines in the consumer terminal, MPD provides direction for its evolution.

\begin{figure}
\centering
\includegraphics[width = \linewidth]{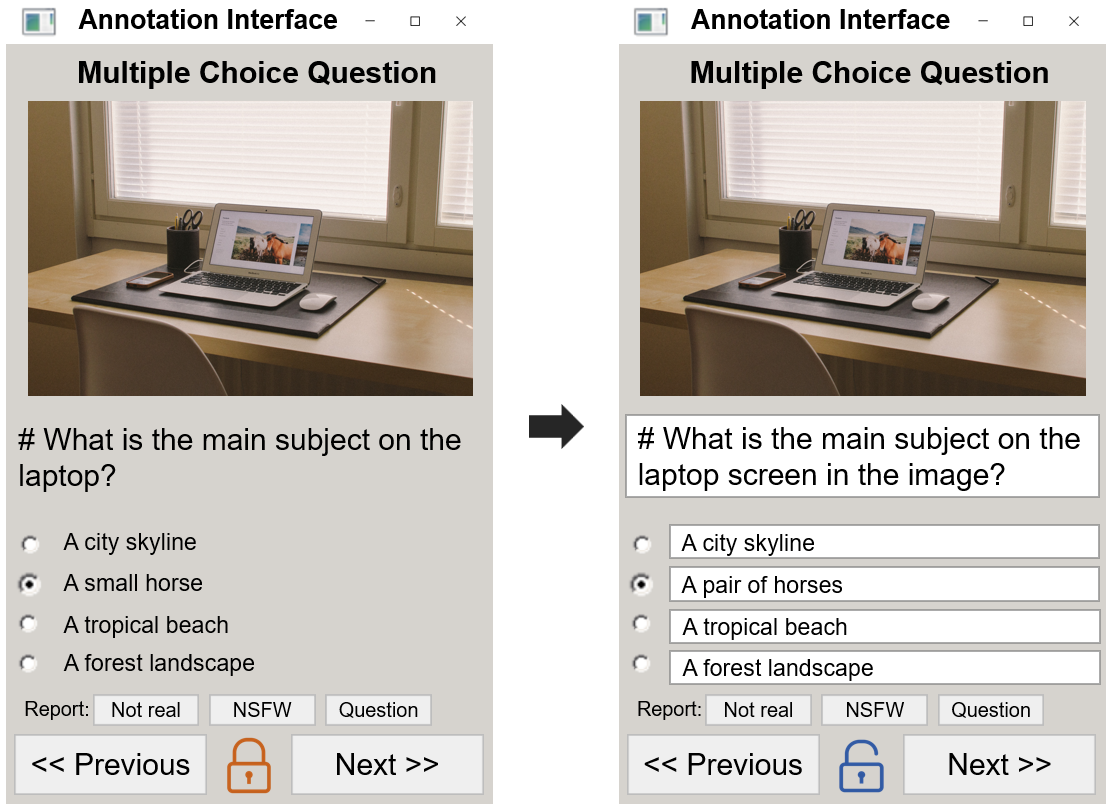}
\caption{Human annotation interface for YoN/MCQ/VQA tasks. Taking MCQ as an example, viewers can answer directly (left) or modify question/choice (right).}
\label{fig:interface}
\end{figure}

\section{Human Expert Labeling}

Seven experts with experience in LMM research project participated in the annotation, targeting 1,000 high-quality reference images in MPD.
Figure \ref{fig:interface} shows the user annotation interface, where subjects will complete interspersed YoN/MCQ/VQA/CAP tasks. Taking MCQ as an example, subjects can make the following decisions:

\begin{itemize}
    \item If they agree with the question design, answer directly and click {\tt Next};
    \item If they find some choice is unclear, click {\tt Unlock} to get permission to re-edit the content;
    \item If they find the question is unclear or too easy, click {\tt Question} after {\tt Unlock}. They may re-design the question-answer pairs and another expert will review whether it is qualified;
    \item If they find the image is {\tt unnatural} or {\tt NSFW}, click the corresponding button to exclude it.
\end{itemize}

The principle of YoN/MCQ/VQA question design is to ask about certain aspects of images as detailed as possible. Do not ask a general question like {\tt What does the image show?}, as this is no different from the CAP task. For the MCQ task, the three incorrect options need to be as long as the correct option and have the same part of speech to ensure challenge. The expected output for YoN and MCQ is an option (for human experts) and the probability of each option (for LMMs). The expected output for VQA is a short phrase less than 5 words, and for CAP it is a text description of 30-40 words. There is no human involvement in the SEG and DET tasks. In particular, the retrieval range for the RET task is 1,000 category labels generated by GPT-4o, those labels have been double-checked by human experts, thus ensuring the reliability of the proposed MPD.

\begin{figure*}
\centering
\includegraphics[width = \linewidth]{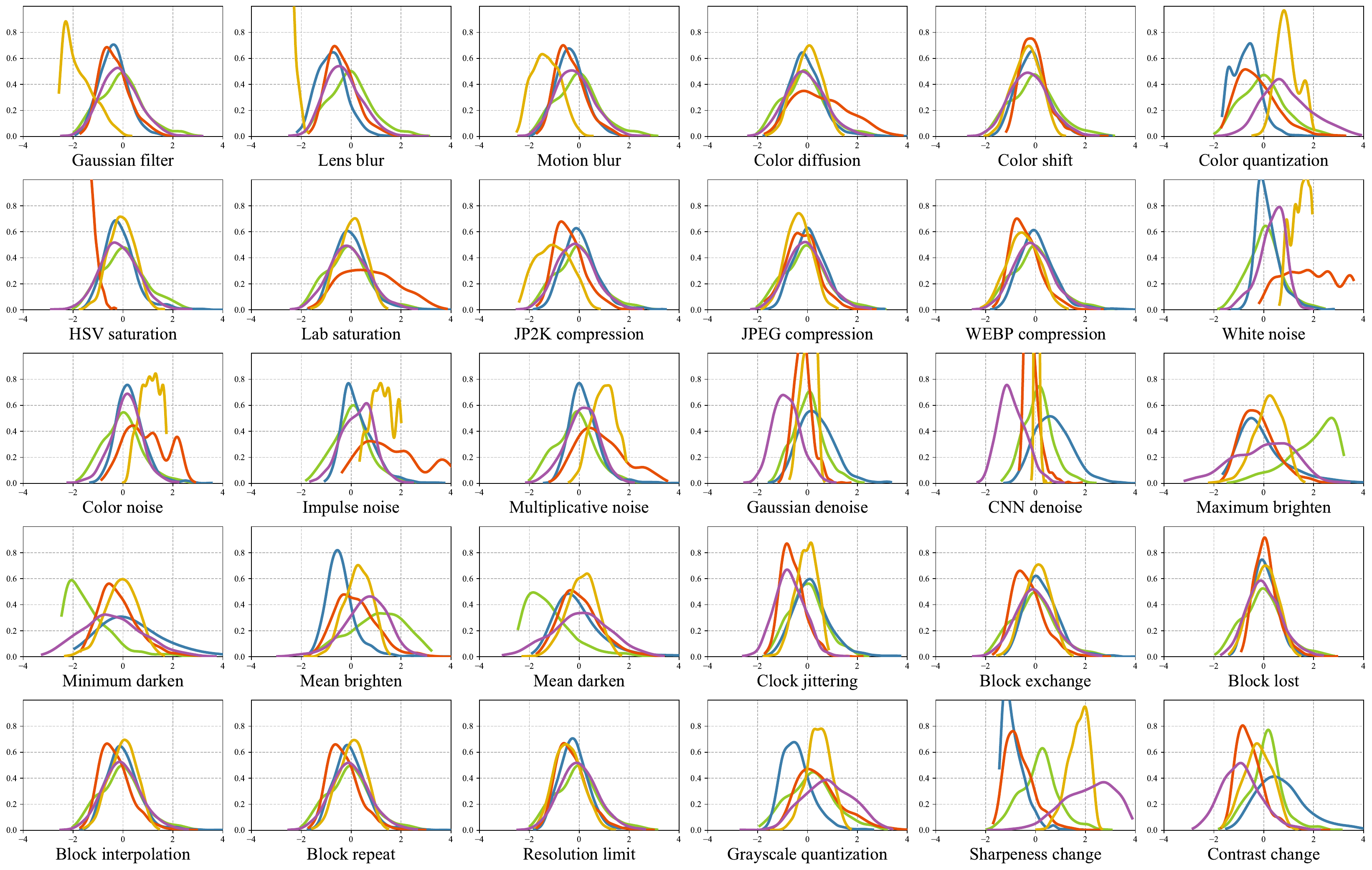}
\vspace{-2mm}
\caption{Low-level feature distribution of MPD, normalized and visualized in 30 corruption subsets. Different colors denote \GroupA{Luminance}, \GroupB{Contrast}, \GroupC{Chrominance}, \GroupD{Blur}, and \GroupE{Spatial Information}.}
\vspace{-2mm}
\label{fig:low-level}
\end{figure*}

\begin{table*}[t]
\centering
    \caption{The definition of all 30 corruptions, and their 5 strength level. For each level, the machine preference score in MPD is shown as mean (std). The validation set is split as \CLA{Mild} and {Severe} distortion.}
    \label{tab:split}
    \vspace{-8pt}
    \renewcommand\arraystretch{1.4}
    \belowrulesep=0pt\aboverulesep=0pt
    \resizebox{\linewidth}{!}{
\begin{tabular}{l|ccccc|l}
\toprule
Corruption                           & Strength 1            & Strength 2            & Strength 3            & Strength 4            & Strength 5 & \multicolumn{1}{c}{Definition}  \\ \midrule
01: Gaussian filter        & \CLA{3.424 (0.203)} & 3.265 (0.201)          & 3.112 (0.233)          & 2.984 (0.255)          & 2.812 (0.284) & Filter with a variable Gaussian kernel \\
02: Lens blur              & 2.857 (0.296)          & 2.779 (0.318)          & 2.363 (0.313)          & 1.999 (0.338)          & 1.685 (0.254) & Filter with a circular kernel\\
03: Motion blur            & 3.300 (0.225)          & 3.074 (0.266)          & 2.893 (0.315)          & 2.738 (0.321)          & 2.488 (0.346) & Filter with a line kernel \\
04: Color diffusion        & \CLA{3.571 (0.204)} & \CLA{3.464 (0.190)} & \CLA{3.374 (0.214)} & 3.299 (0.222)          & 3.197 (0.242) & Gaussian blur the color channels in the Lab color-space \\
05: Color shift            & 3.144 (0.248)          & 3.116 (0.230)          & 3.118 (0.260)          & 3.087 (0.250)          & 3.115 (0.248) &  Randomly translate the green channel\\
06: Color quantization     & 3.207 (0.260)          & 3.055 (0.284)          & 2.904 (0.276)          & 2.807 (0.300)          & 2.503 (0.347) & Using minimum variance quantization in limited color  \\
07: HSV saturation         & 2.989 (0.367)          & 2.975 (0.351)          & 2.933 (0.341)          & 2.973 (0.334)          & 2.946 (0.357) & Multiply the saturation channel in the HSV color-space \\
08: Lab saturation         & \CLA{4.501 (0.232)} & \CLA{3.673 (0.180)} & \CLA{3.480 (0.229)} & \CLA{3.376 (0.220)} & 3.311 (0.259) & Multiply the saturation channel in the Lab color-space \\
09: JP2K compression       & 3.166 (0.269)          & 2.942 (0.303)          & 2.831 (0.301)          & 2.588 (0.377)          & 2.441 (0.367) & A standard image codec \\
10: JPEG compression       & 2.875 (0.269)          & 2.600 (0.313)          & 2.446 (0.291)          & 2.427 (0.269)          & 2.402 (0.304) & The most widely used image codec \\
11: WEBP compression       & 3.053 (0.231)          & 3.034 (0.231)          & 3.052 (0.239)          & 2.985 (0.239)          & 2.918 (0.267) & Codec with the best comprehensive performance \\
12: White noise            & 3.056 (0.242)          & 2.842 (0.243)          & 2.666 (0.277)          & 2.423 (0.325)          & 2.106 (0.332) & Gaussian white noise to RGB channels\\
13: Color noise            & 3.255 (0.207)          & 3.074 (0.232)          & 2.938 (0.248)          & 2.760 (0.267)          & 2.457 (0.329) & Gaussian white noise to the YCbCr converted image\\
14: Impulse noise          & 3.193 (0.214)          & 3.077 (0.232)          & 2.911 (0.245)          & 2.650 (0.283)          & 2.120 (0.322) & Salt and pepper noise to RGB channels\\
15: Multiplicative noise   & \CLA{3.368 (0.228)} & 3.176 (0.246)          & 3.034 (0.259)          & 2.894 (0.266)          & 2.617 (0.293) & Speckle noise to RGB channels \\
16: Gaussian denoise       & 2.807 (0.284)          & 2.655 (0.316)          & 2.602 (0.275)          & 2.445 (0.342)          & 2.345 (0.329) & New noise introduced by Gaussian filter denoise\\
17: CNN denoise            & 2.740 (0.292)          & 2.586 (0.278)          & 2.514 (0.298)          & 2.403 (0.338)          & 2.286 (0.321) & AI-artifacts introduced by neural network denoise \\
18: Maximum brighten       & \CLA{3.395 (0.193)} & 3.152 (0.297)          & 2.964 (0.278)          & 2.782 (0.309)          & 2.610 (0.360) & Keep the maximum pixel value and increase others\\
19: Minimum darken         & \CLA{3.522 (0.193)} & 3.251 (0.275)          & 3.018 (0.283)          & 2.706 (0.374)          & 2.485 (0.398) & Keep the minimum pixel value and increase others\\
20: Mean brighten          & \CLA{3.610 (0.216)} & \CLA{3.482 (0.243)} & \CLA{3.370 (0.271)} & 3.204 (0.258)          & 3.057 (0.294) & Increase all pixel values and truncate to the original range\\
21: Mean darken            & \CLA{3.338 (0.274)} & 3.192 (0.307)          & 3.024 (0.363)          & 2.860 (0.369)          & 2.687 (0.385) & Decrease all pixel values and truncate to the original range\\
22: Clock jittering        & \CLA{3.472 (0.240)} & 3.102 (0.301)          & 2.796 (0.311)          & 2.472 (0.376)          & 2.064 (0.383) & Warping each pixel with random small offsets\\
23: Block exchange         & 3.161 (0.237)          & 3.083 (0.216)          & 2.958 (0.204)          & 2.874 (0.209)          & 2.729 (0.223) & The order of two Blocks is wrong \\
24: Block lost             & 3.248 (0.223)          & 3.193 (0.208)          & 3.091 (0.194)          & 2.993 (0.193)          & 2.880 (0.198) & A Block is lost and turned into random pixels\\
25: Block interpolation    & \CLA{3.547 (0.202)} & \CLA{3.473 (0.202)} & \CLA{3.379 (0.188)} & 3.290 (0.187)          & 3.213 (0.193) & A Block is lost and recovered by surrounding pixels\\
26: Block repeat           & \CLA{3.605 (0.219)} & \CLA{3.554 (0.217)} & \CLA{3.433 (0.236)} & \CLA{3.388 (0.226)} & 3.228 (0.244) &  A Block is convoluted twice on the original position\\
27: Resolution limit       & 3.328 (0.242)          & 3.108 (0.290)          & 2.784 (0.325)          & 2.562 (0.344)          & 2.359 (0.344) & Downsize the image and upsize it back\\
28: Grayscale quantization & 3.332 (0.252)          & 3.248 (0.269)          & 3.148 (0.246)          & 3.008 (0.306)          & 2.840 (0.281) & Quantize image values using Otsus thresholds\\
29: Sharpness change      & 3.110 (0.218)          & 2.909 (0.250)          & 2.763 (0.255)          & 2.589 (0.287)          & 2.492 (0.306) & Over-sharpen image using unsharp masking\\
30: Contrast change        & \CLA{3.703 (0.171)} & \CLA{3.606 (0.184)} & 3.258 (0.221)          & 3.159 (0.248)          & 3.037 (0.291) & Non-linearly change RGB values by Sigmoid\\
\bottomrule
\end{tabular}}
\vspace{4mm}
\end{table*}

\begin{table*}[t]
\centering
    \caption{Using IQA metrics to predict machine preference, including baseline, FR, and NR algorithm, human visual system-based metrics are marked in {\faCheckCircle}. The first row represents two distortion intensities and three image content subsets, and the second row represents five scoring dimensions. [Keys: \CLB{Best}; \CLA{Second best}; \textbf{Baseline (BL)}; \colorbox{gray!20}{Inferior} to the baseline.]}
    \label{tab:str}
    \vspace{-8pt}
    \renewcommand\arraystretch{1.3}
    \belowrulesep=0pt\aboverulesep=0pt
    \resizebox{\linewidth}{!}{
\begin{tabular}{c|l|c|ccc|ccc|ccc|ccc|ccc}
    \toprule
                           & \multicolumn{1}{c|}{}                         & \multicolumn{1}{c|}{}                      & \multicolumn{3}{c|}{Strength 1}                                                                                    & \multicolumn{3}{c|}{Strength 2}                                                                                                              & \multicolumn{3}{c|}{Strength 3}                                                                                                  & \multicolumn{3}{c|}{Strength 4}                                                                                                  & \multicolumn{3}{c}{Strength 5}                                                                                                 \\ \cdashline{4-18} 
    \multirow{-2}{*}{Type}     & \multicolumn{1}{c|}{\multirow{-2}{*}{Metric}} & \multicolumn{1}{c|}{\multirow{-2}{*}{HVS}} & SRCC$\uparrow$                                  & KRCC$\uparrow$                                  & PLCC$\uparrow$                                   & SRCC$\uparrow$                                  & KRCC$\uparrow$                                  & PLCC$\uparrow$                                                           & SRCC$\uparrow$                                  & KRCC$\uparrow$                                  & PLCC$\uparrow$                                   & SRCC$\uparrow$                                  & KRCC$\uparrow$                                  & PLCC$\uparrow$                                   & SRCC$\uparrow$                                  & KRCC$\uparrow$                                  & PLCC$\uparrow$                                   \\ \midrule
\multirow{2}{*}{BL} & PSNR   &  & 0.3348          & 0.2400          & \textbf{0.6169} & 0.2269          & 0.1497          & 0.2450          & 0.2900          & 0.1958          & 0.3275          & 0.2584          & 0.1733          & 0.2632          & 0.3140          & 0.2115          & 0.3406          \\
                          & SSIM  &   & \textbf{0.5455} & \textbf{0.3720} & 0.4569          & \textbf{0.5176} & \textbf{0.3571} & \textbf{0.4926} & \textbf{0.4963} & \textbf{0.3395} & \textbf{0.4423} & \textbf{0.4515} & \textbf{0.3058} & \textbf{0.4056} & \textbf{0.5708} & \textbf{0.4008} & \textbf{0.5507} \\ \cdashline{1-18}

\multirow{6}{*}{FR}       & AHIQ  &   & 0.6410   & 0.4625   & \colorbox{gray!20}{0.6130}   & 0.7514   & 0.5639   & 0.7986   & \CLA{0.7618}   & \CLA{0.5793}   & \CLB{0.7820}   & 0.7262   & 0.5340   & 0.7273   & \CLB{0.8533}   & \CLB{0.6721}   & \CLB{0.8760}   \\
                          & CKDN  & \faCheckCircle  & 0.5770   & 0.4030   & \colorbox{gray!20}{0.4431}   & 0.5966   & 0.4256   & 0.5923   & 0.6285   & 0.4406   & 0.6165   & 0.6347   & 0.4285   & 0.5953   & 0.7538   & 0.5290   & 0.7472   \\
                          & DISTS  &  & 0.6813   & 0.4856   & 0.6576   & 0.7013   & 0.5183   & 0.7254   & 0.6569   & 0.4810   & 0.6520   & 0.6471   & 0.4489   & 0.6411   & 0.8324   & 0.6481   & 0.8227   \\
                          & LPIPS  & \faCheckCircle & \colorbox{gray!20}{0.5238}   & \colorbox{gray!20}{0.3544}   & \colorbox{gray!20}{0.3762}   & 0.6266   & 0.4431   & 0.5600   & 0.5375   & 0.3846   & 0.4647   & \colorbox{gray!20}{0.4229}   & \colorbox{gray!20}{0.2909}   & \colorbox{gray!20}{0.3853}   & \colorbox{gray!20}{0.5138}   & \colorbox{gray!20}{0.3720}   & \colorbox{gray!20}{0.4716}   \\
                          & PieAPP   & \faCheckCircle & \colorbox{gray!20}{0.5145}   & \colorbox{gray!20}{0.1204}   & \colorbox{gray!20}{0.3484}   & \colorbox{gray!20}{0.4180}   & \colorbox{gray!20}{0.0270}   & \colorbox{gray!20}{0.2031}   & \colorbox{gray!20}{0.3717}   & \colorbox{gray!20}{0.1533}   & \colorbox{gray!20}{0.1943}   & \colorbox{gray!20}{0.3241}   & \colorbox{gray!20}{0.0855}   & \colorbox{gray!20}{0.1411}   & 0.5709   & \colorbox{gray!20}{0.2428}   & \colorbox{gray!20}{0.3458}   \\
                          & TOPIQ-FR & \faCheckCircle & 0.6237   & 0.4358   & 0.6481   & 0.7066   & 0.5166   & 0.7149   & 0.6827   & 0.5006   & 0.6711   & 0.6658   & 0.4594   & 0.6462   & 0.7804   & 0.5842   & 0.7748   \\ \cdashline{1-18}
\multirow{6}{*}{NR}       & ARNIQA  & & \CLB{0.7773}   & \CLA{0.5424}   & \CLA{0.6305}   & \CLB{0.8197}   & \CLB{0.6240}   & \CLB{0.8339}   & \CLB{0.7953}   & \CLB{0.5927}   & \CLA{0.7416}   & \CLB{0.7731}   & \CLA{0.5664}   & \CLA{0.7331}   & \CLA{0.8503}   & 0.6449   & 0.8353   \\
                          & CLIPIQA & \faCheckCircle & \colorbox{gray!20}{0.4365}   & \colorbox{gray!20}{0.2928}   & \colorbox{gray!20}{0.3682}   & 0.5556   & 0.3952   & 0.5918   & \colorbox{gray!20}{0.4809}   & \colorbox{gray!20}{0.3162}   & 0.4560   & 0.6352   & 0.4360   & 0.5570   & 0.6970   & 0.5023   & 0.6234   \\
                          & DBCNN    & \faCheckCircle & \colorbox{gray!20}{0.4863}   & \colorbox{gray!20}{0.3355}   & \colorbox{gray!20}{0.4736}   & 0.5338   & 0.3690   & 0.5566   & \colorbox{gray!20}{0.4682}   & \colorbox{gray!20}{0.3095}   & 0.4633   & \colorbox{gray!20}{0.4420}   & \colorbox{gray!20}{0.2921}   & 0.4226   & 0.6430   & 0.4522   & 0.6315   \\
                          & HyperIQA & & \CLA{0.7671}   & \CLB{0.5678}   & \CLB{0.6647}   & \CLA{0.8046}   & \CLA{0.6097}   & \CLA{0.8211}   & 0.7406   & 0.5566   & 0.7543   & 0.7388   & 0.5469   & 0.7473   & 0.8384   & \CLA{0.6569}   & \CLA{0.8604}   \\
                          & NIMA   &  & 0.5597   & 0.3972   & \colorbox{gray!20}{0.5243}   & 0.7736   & 0.5776   & 0.7767   & 0.7461   & 0.5532   & 0.7831   & \CLA{0.7621}   & \CLB{0.5697}   & \CLB{0.7607}   & 0.8208   & 0.6351   & 0.8488   \\
                          & TOPIQ-NR & \faCheckCircle & 0.5650   & 0.3893   & \colorbox{gray!20}{0.5566}   & 0.6056   & 0.4331   & 0.6326   & 0.5852   & 0.4109   & 0.5894   & 0.6407   & 0.4417   & 0.6213   & 0.7219   & 0.4778   & 0.6801  
   \\ \bottomrule     
    \end{tabular}}
    \vspace{4mm}
\end{table*}

\section{Image Feature Distribution}

Figure \ref{fig:low-level} shows the distribution of low-level features of all instances of MPD. After overall regularization, 30 types of corruption are grouped and displayed. The features considered include Luminance, Contrast, Chrominance, Blur, and Spatial Information, and their calculation methods can be found in \cite{dataset:agiqa-3k}. There are significant differences in these low-level attributes for different corruptions. For example, in the first three blurry cases, the blur curve is left-biased and becomes right-biased after sharpening. In general, similar corruption categories will lead to similar results (such as five noise-related and four block-related). Two of the denoises have the sharpest distributions; Color quantization, Grayscale quantization, Sharpness change, and Contrast change are the most irregular. These findings deserve further exploration.

\section{Corruption Level}

\subsection{Mild\&Severe Distortion Split}
\label{sec:split}

Due to the different perception mechanisms of MVS and HVS, it is unreasonable to directly regard Strength 1 as mild, because the machine may be very sensitive to a certain distortion but very robust to another. Therefore, we measured the MPD scores of 30 corruptions and 5 strengths, as shown in Table \ref{tab:split}. Overall, as the strength increases, the MOS given by the machine gradually decreases; at the same time, there are huge differences in each corruption itself, which confirms the previous point of view that each corruption needs to be considered separately when dividing. Here, we use 3.333 as the threshold and select a part as the mild validation set in the main text. The specific definition of each corruption is explained in Table \ref{tab:split} as a reference.

\subsection{IQA Evaluation on Different Strength}

Table \ref{tab:str} simply shows the performance of IQA algorithms at different Strength levels. Relatively, the performance of ARNIQA and HyperIQA is more satisfactory. However, as shown in each column of Table \ref{tab:split}, there are huge differences in machine preferences under the same Strength. Therefore, this is not a challenging task. The performance of the IQA model can only be truly reflected b  the MVS-based division in the section \ref{sec:split}, rather than the HVS-based division here.

\begin{figure*}
\centering
\begin{minipage}[]{0.24\linewidth}
  \centering
  \centerline{\includegraphics[width = \textwidth]{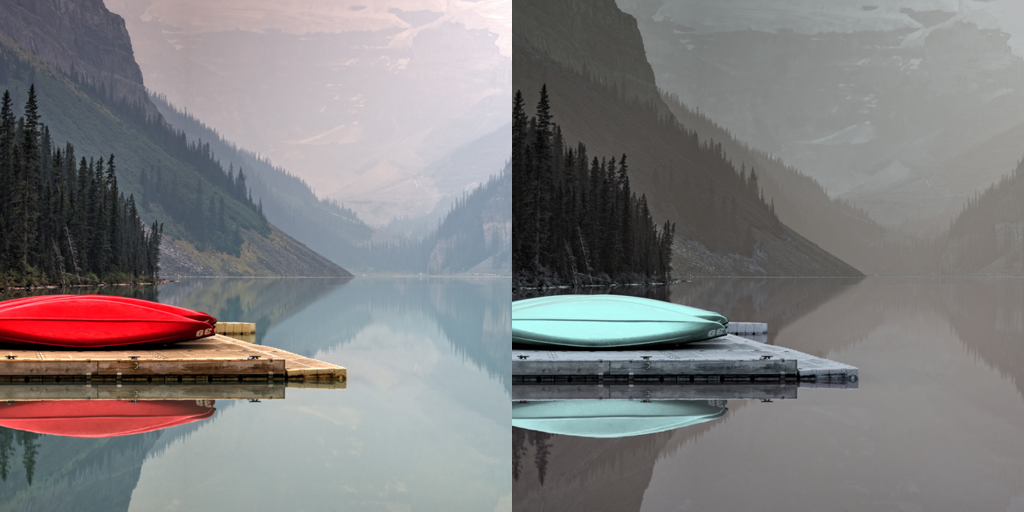}}
  \centerline{MOS=1.7245}\medskip
\end{minipage}
\begin{minipage}[]{0.24\linewidth}
  \centering
  \centerline{\includegraphics[width = \textwidth]{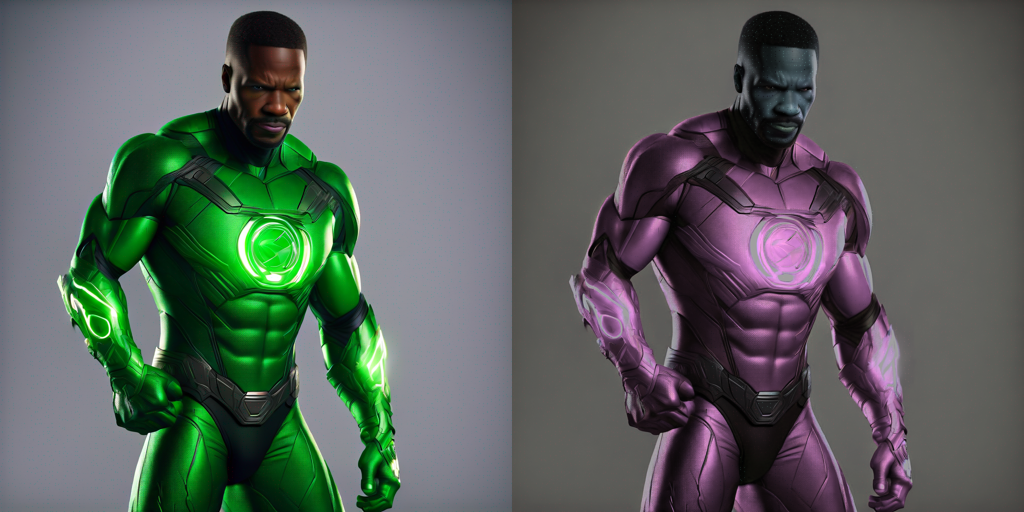}}
  \centerline{MOS=1.8017}\medskip
\end{minipage}
\begin{minipage}[]{0.24\linewidth}
  \centering
  \centerline{\includegraphics[width = \textwidth]{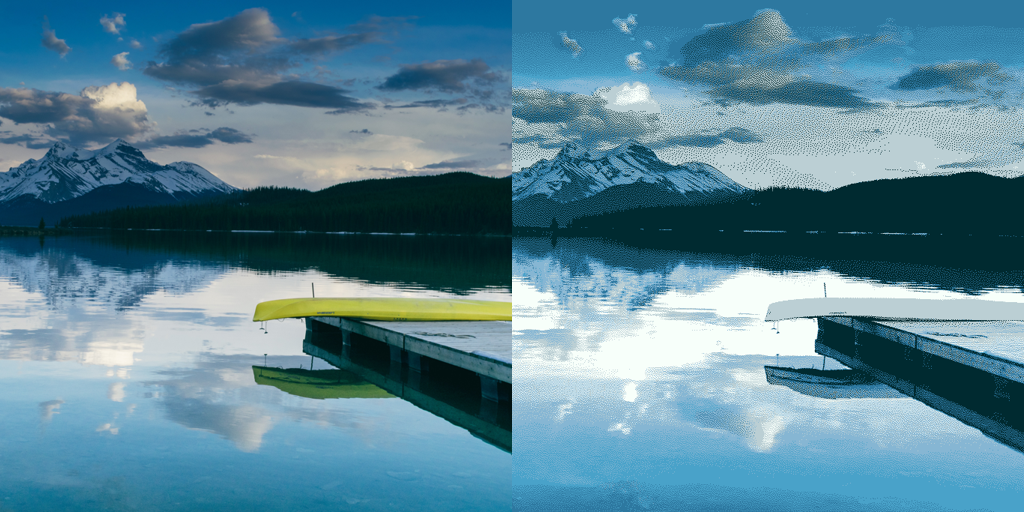}}
  \centerline{MOS=1.8143}\medskip
\end{minipage}
\begin{minipage}[]{0.24\linewidth}
  \centering
  \centerline{\includegraphics[width = \textwidth]{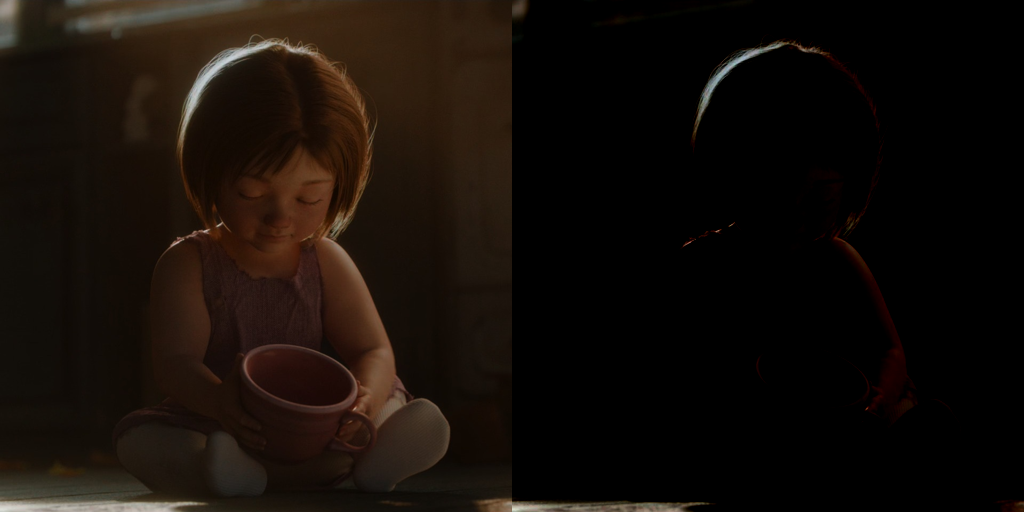}}
  \centerline{MOS=1.8407}\medskip
\end{minipage}

\begin{minipage}[]{0.24\linewidth}
  \centering
  \centerline{\includegraphics[width = \textwidth]{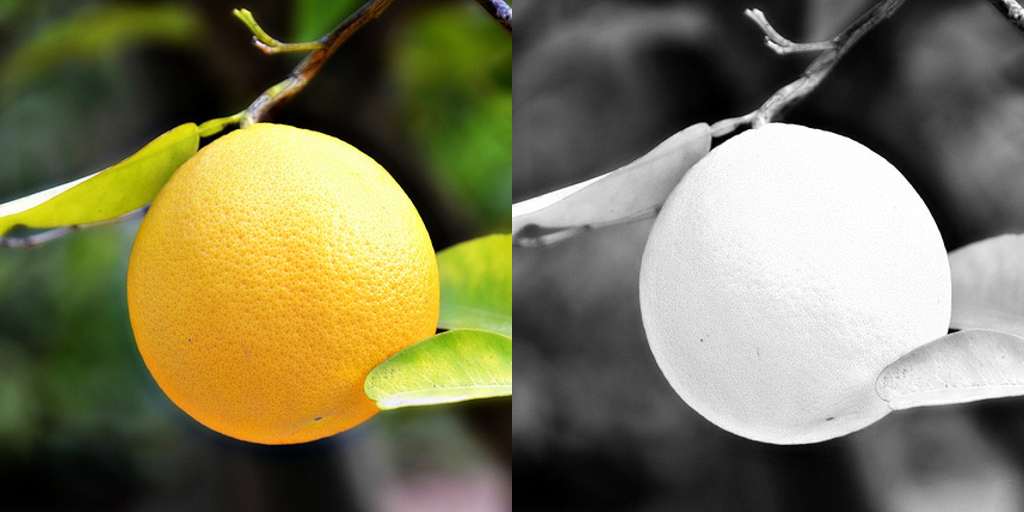}}
  \centerline{MOS=1.8901}\medskip
\end{minipage}
\begin{minipage}[]{0.24\linewidth}
  \centering
  \centerline{\includegraphics[width = \textwidth]{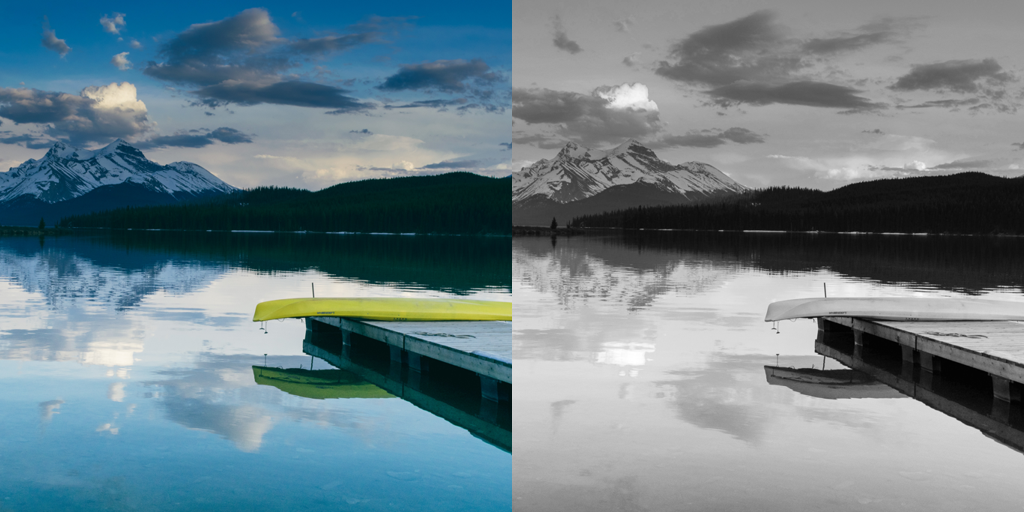}}
  \centerline{MOS=1.9289}\medskip
\end{minipage}
\begin{minipage}[]{0.24\linewidth}
  \centering
  \centerline{\includegraphics[width = \textwidth]{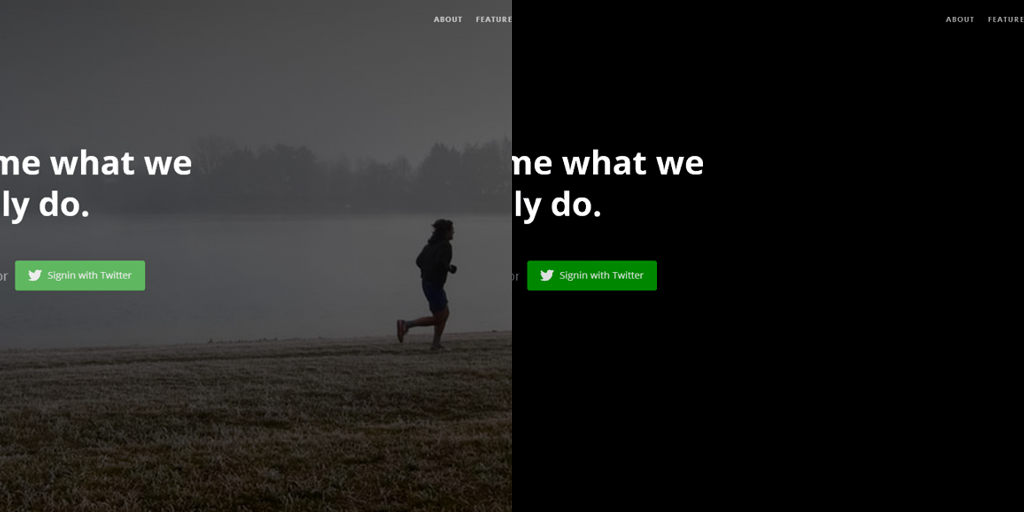}}
  \centerline{MOS=1.9417}\medskip
\end{minipage}
\begin{minipage}[]{0.24\linewidth}
  \centering
  \centerline{\includegraphics[width = \textwidth]{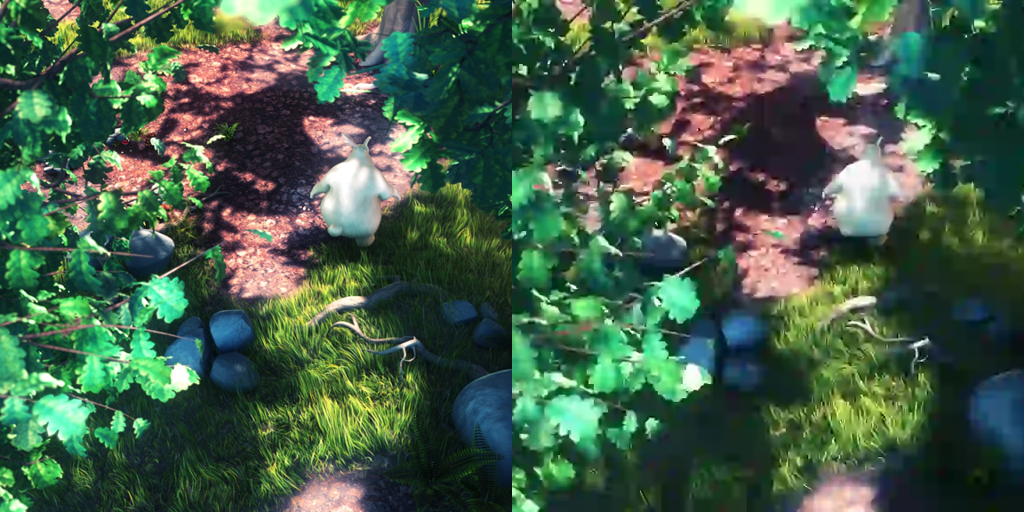}}
  \centerline{MOS=1.9426}\medskip
\end{minipage}

\begin{minipage}[]{0.24\linewidth}
  \centering
  \centerline{\includegraphics[width = \textwidth]{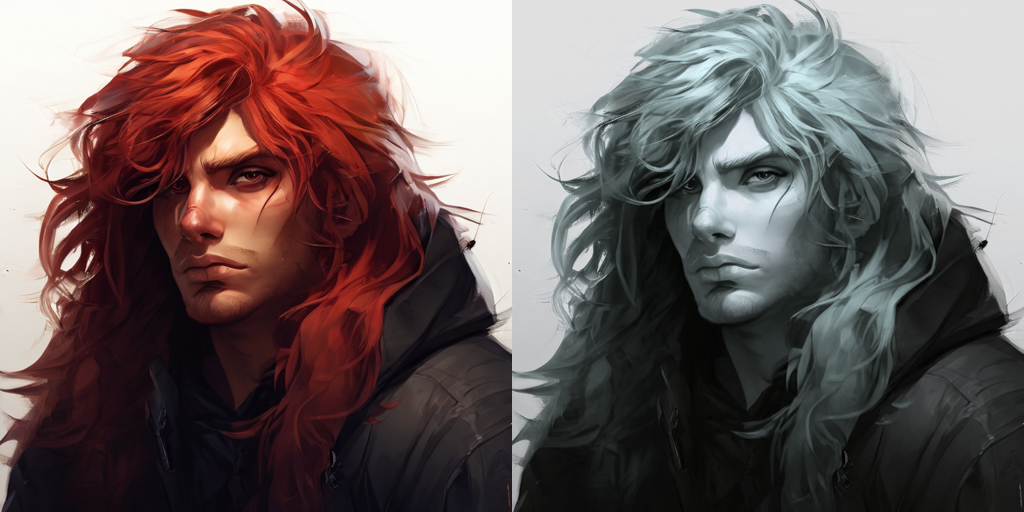}}
  \centerline{MOS=1.9445}\medskip
\end{minipage}
\vspace{-2mm}
\begin{minipage}[]{0.24\linewidth}
  \centering
  \centerline{\includegraphics[width = \textwidth]{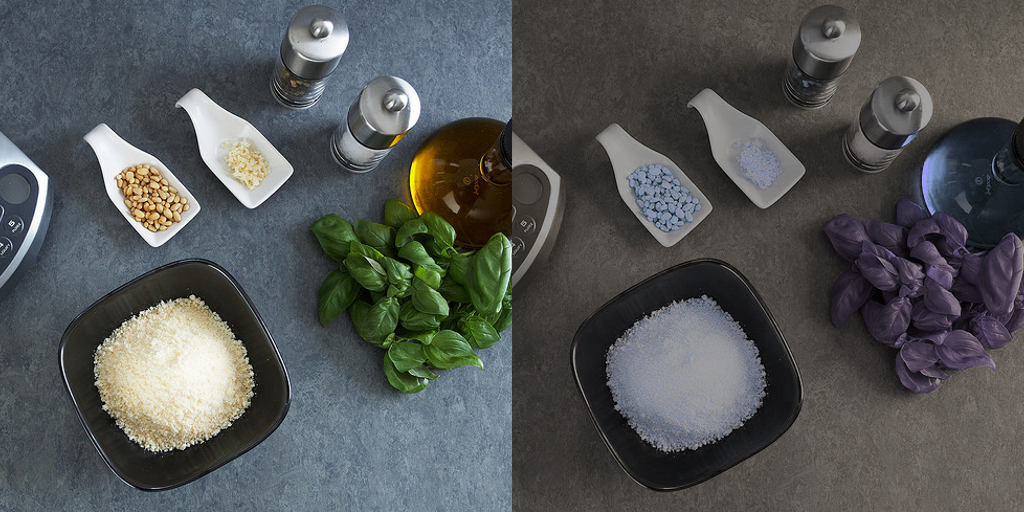}}
  \centerline{MOS=2.0181}\medskip
\end{minipage}
\vspace{-2mm}
\begin{minipage}[]{0.24\linewidth}
  \centering
  \centerline{\includegraphics[width = \textwidth]{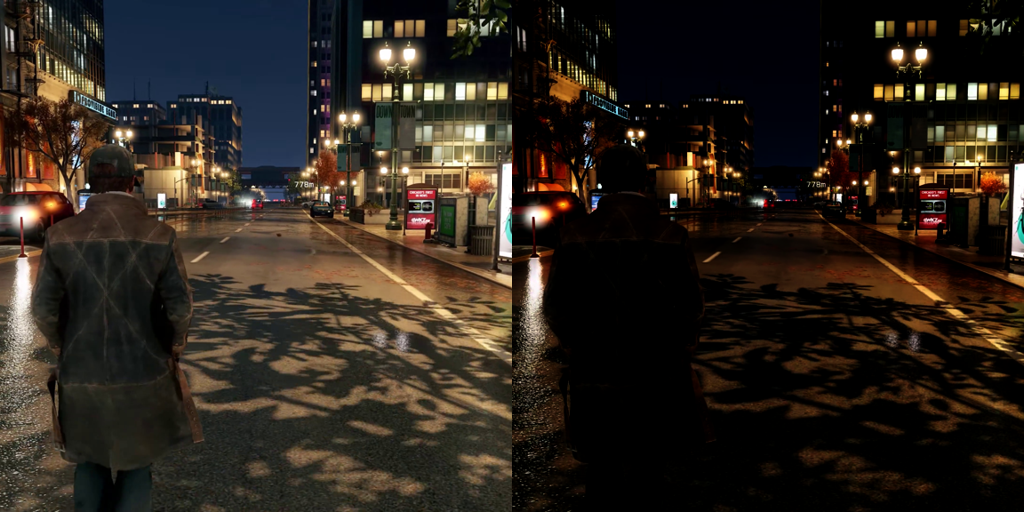}}
  \centerline{MOS=2.1449}\medskip
\end{minipage}
\vspace{-2mm}
\begin{minipage}[]{0.24\linewidth}
  \centering
  \centerline{\includegraphics[width = \textwidth]{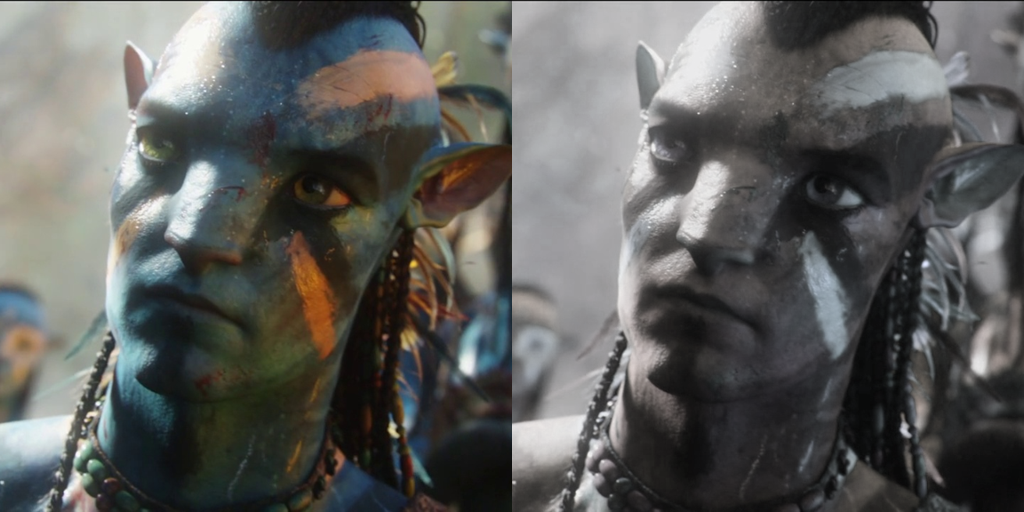}}
  \centerline{MOS=2.1697}\medskip
\end{minipage}
\vspace{-2mm}

\vspace{6mm}
\caption{Example of low-quality images for machine preference. (Left: Reference; Right: Distorted, center-cropped as 512$\times$512)}
\label{fig:low}
\end{figure*}

\begin{figure*}
\centering
\begin{minipage}[]{0.24\linewidth}
  \centering
  \centerline{\includegraphics[width = \textwidth]{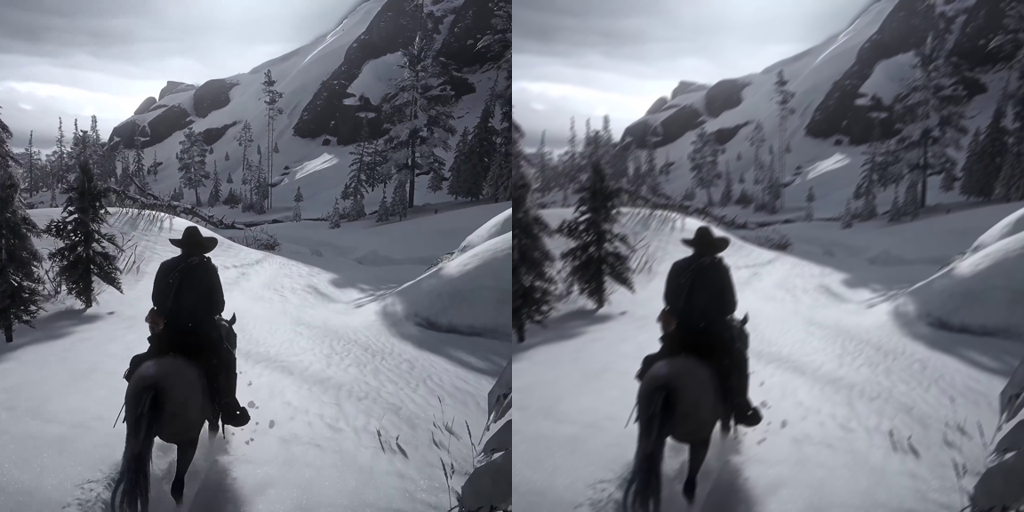}}
  \centerline{MOS=3.8084}\medskip
\end{minipage}
\begin{minipage}[]{0.24\linewidth}
  \centering
  \centerline{\includegraphics[width = \textwidth]{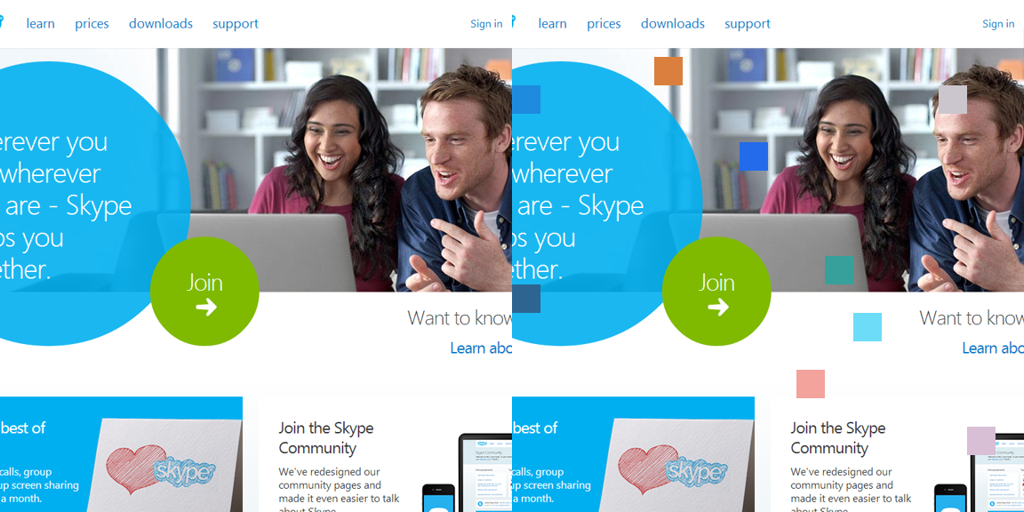}}
  \centerline{MOS=3.8093}\medskip
\end{minipage}
\begin{minipage}[]{0.24\linewidth}
  \centering
  \centerline{\includegraphics[width = \textwidth]{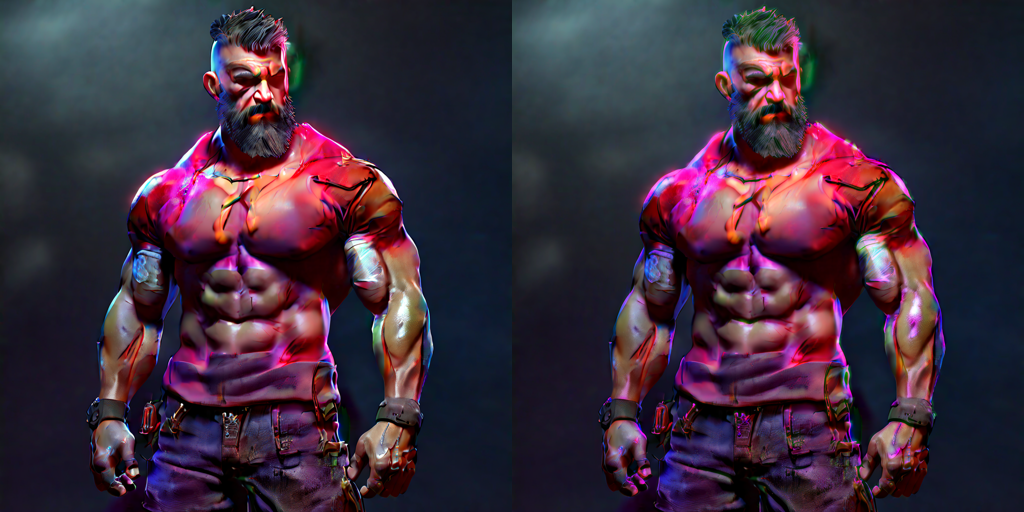}}
  \centerline{MOS=3.8096}\medskip
\end{minipage}
\begin{minipage}[]{0.24\linewidth}
  \centering
  \centerline{\includegraphics[width = \textwidth]{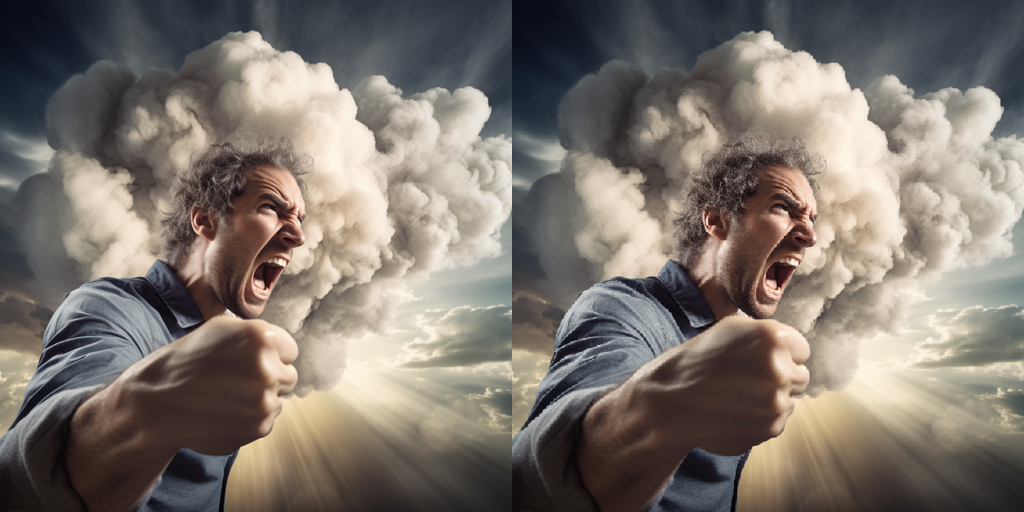}}
  \centerline{MOS=3.8259}\medskip
\end{minipage}

\begin{minipage}[]{0.24\linewidth}
  \centering
  \centerline{\includegraphics[width = \textwidth]{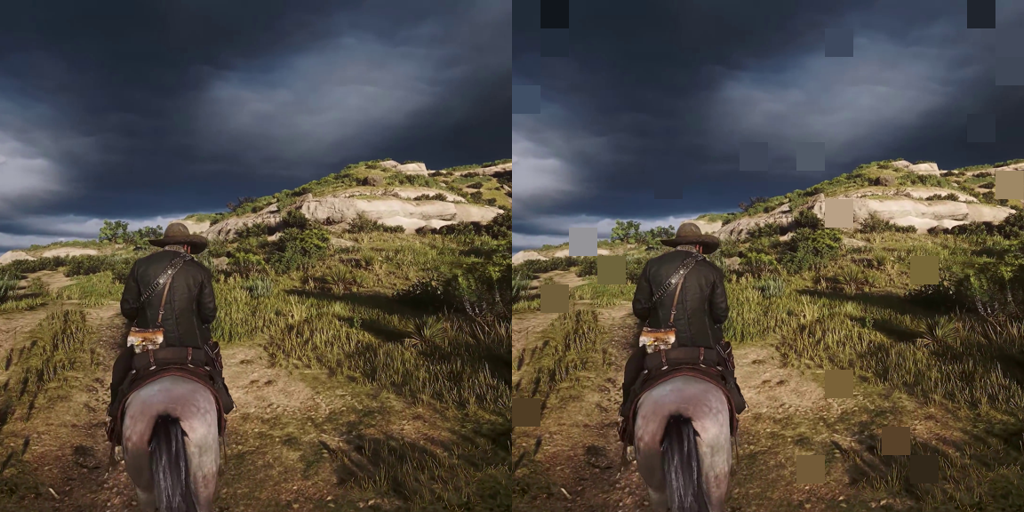}}
  \centerline{MOS=3.8292}\medskip
\end{minipage}
\begin{minipage}[]{0.24\linewidth}
  \centering
  \centerline{\includegraphics[width = \textwidth]{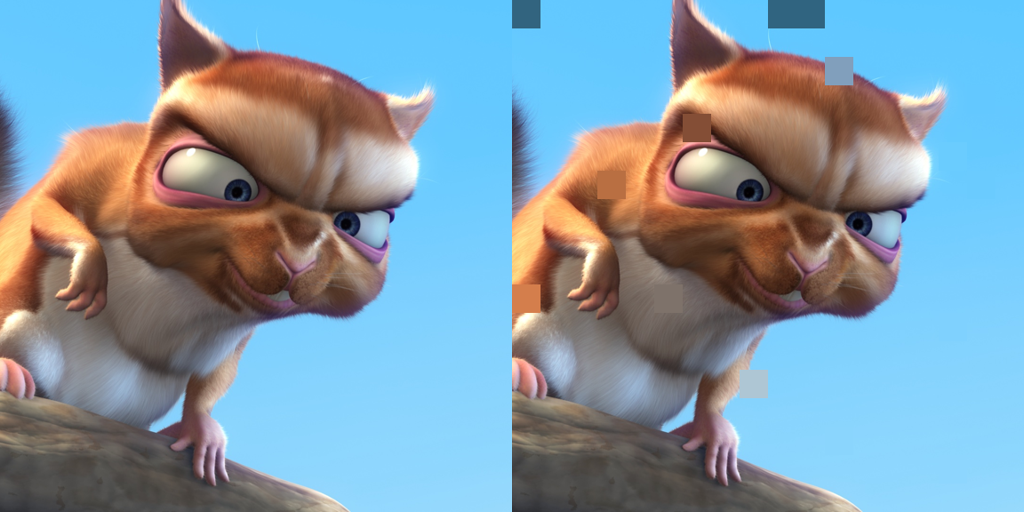}}
  \centerline{MOS=3.8432}\medskip
\end{minipage}
\begin{minipage}[]{0.24\linewidth}
  \centering
  \centerline{\includegraphics[width = \textwidth]{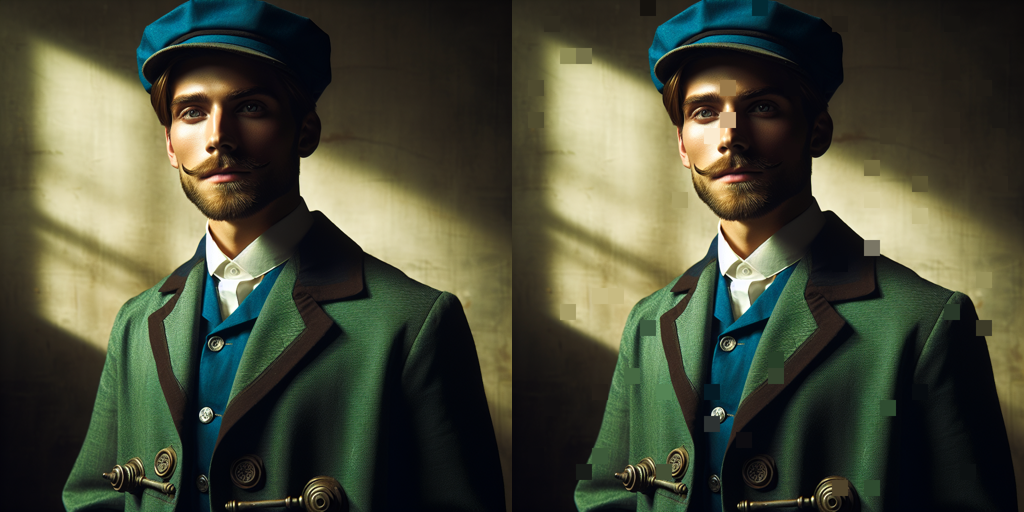}}
  \centerline{MOS=3.8462}\medskip
\end{minipage}
\begin{minipage}[]{0.24\linewidth}
  \centering
  \centerline{\includegraphics[width = \textwidth]{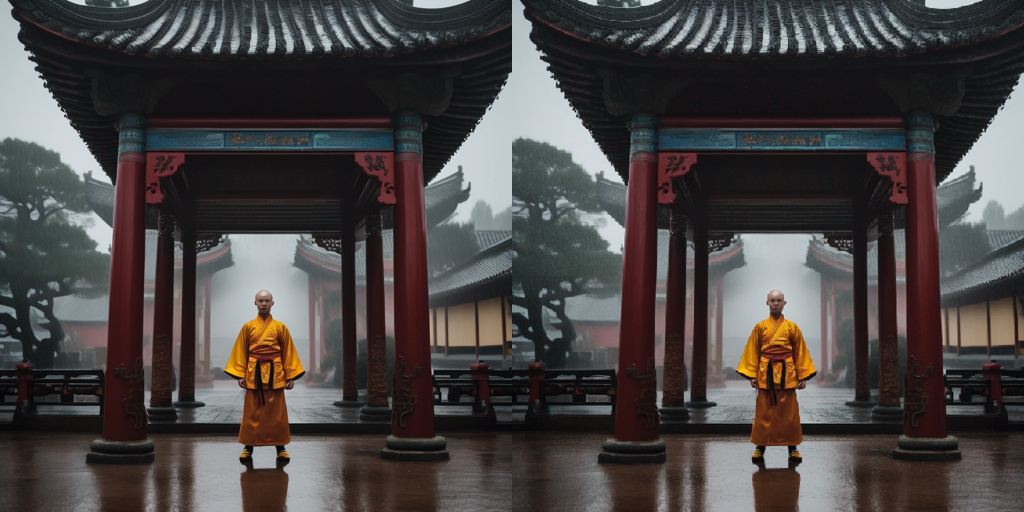}}
  \centerline{MOS=3.8551}\medskip
\end{minipage}

\begin{minipage}[]{0.24\linewidth}
  \centering
  \centerline{\includegraphics[width = \textwidth]{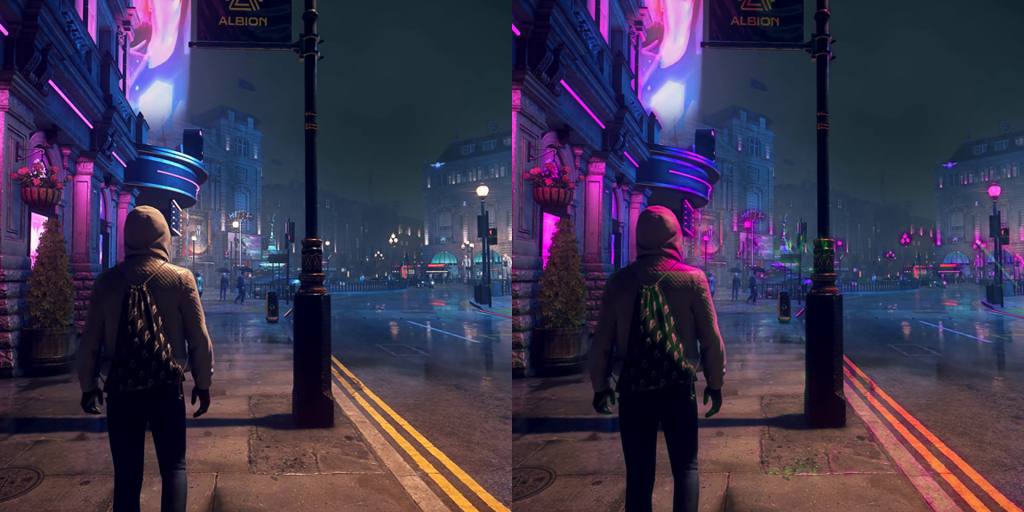}}
  \centerline{MOS=3.9073}\medskip
\end{minipage}
\vspace{-2mm}
\begin{minipage}[]{0.24\linewidth}
  \centering
  \centerline{\includegraphics[width = \textwidth]{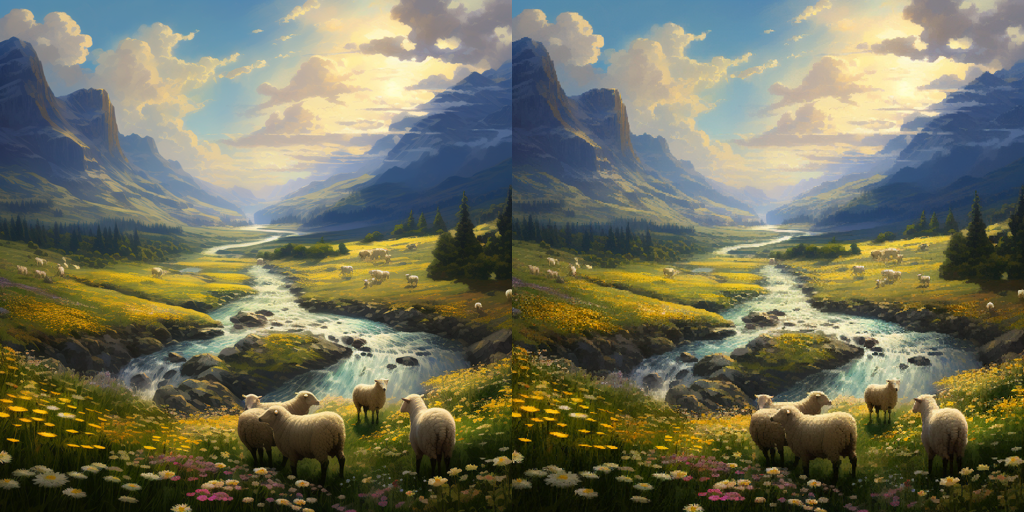}}
  \centerline{MOS=4.0054}\medskip
\end{minipage}
\vspace{-2mm}
\begin{minipage}[]{0.24\linewidth}
  \centering
  \centerline{\includegraphics[width = \textwidth]{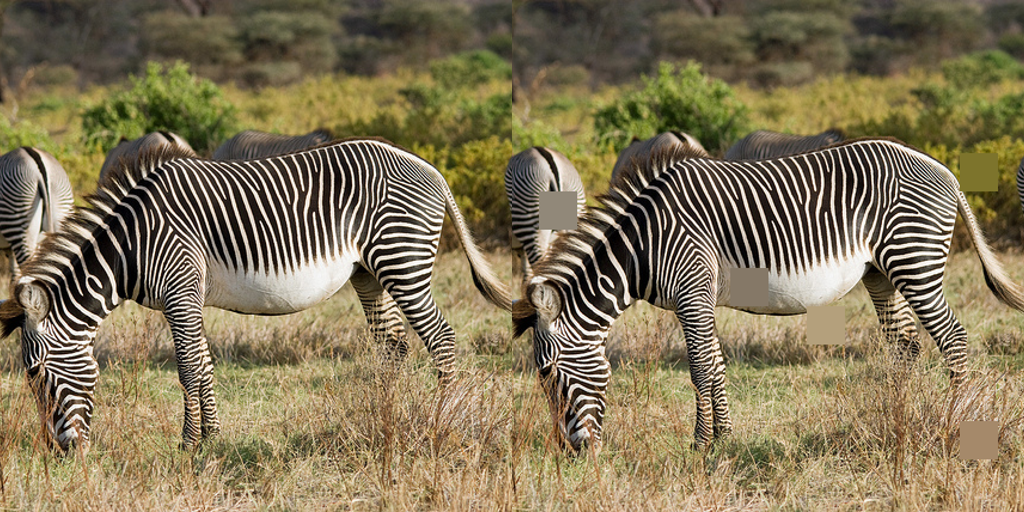}}
  \centerline{MOS=4.0424}\medskip
\end{minipage}
\vspace{-2mm}
\begin{minipage}[]{0.24\linewidth}
  \centering
  \centerline{\includegraphics[width = \textwidth]{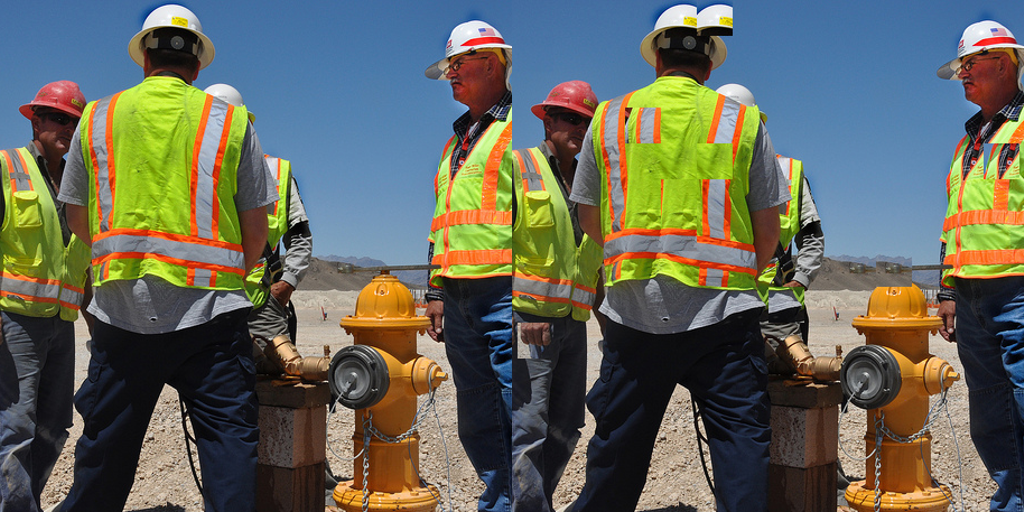}}
  \centerline{MOS=4.1826}\medskip
\end{minipage}
\vspace{-2mm}

\vspace{6mm}
\caption{Example of high-quality images for machine preference. (Left: Reference; Right: Distorted, center-cropped as 512$\times$512)}
\label{fig:high}
\end{figure*}

\subsection{Corruption Type\&Level Example}

\begin{figure*}
\centering
\includegraphics[width = \linewidth]{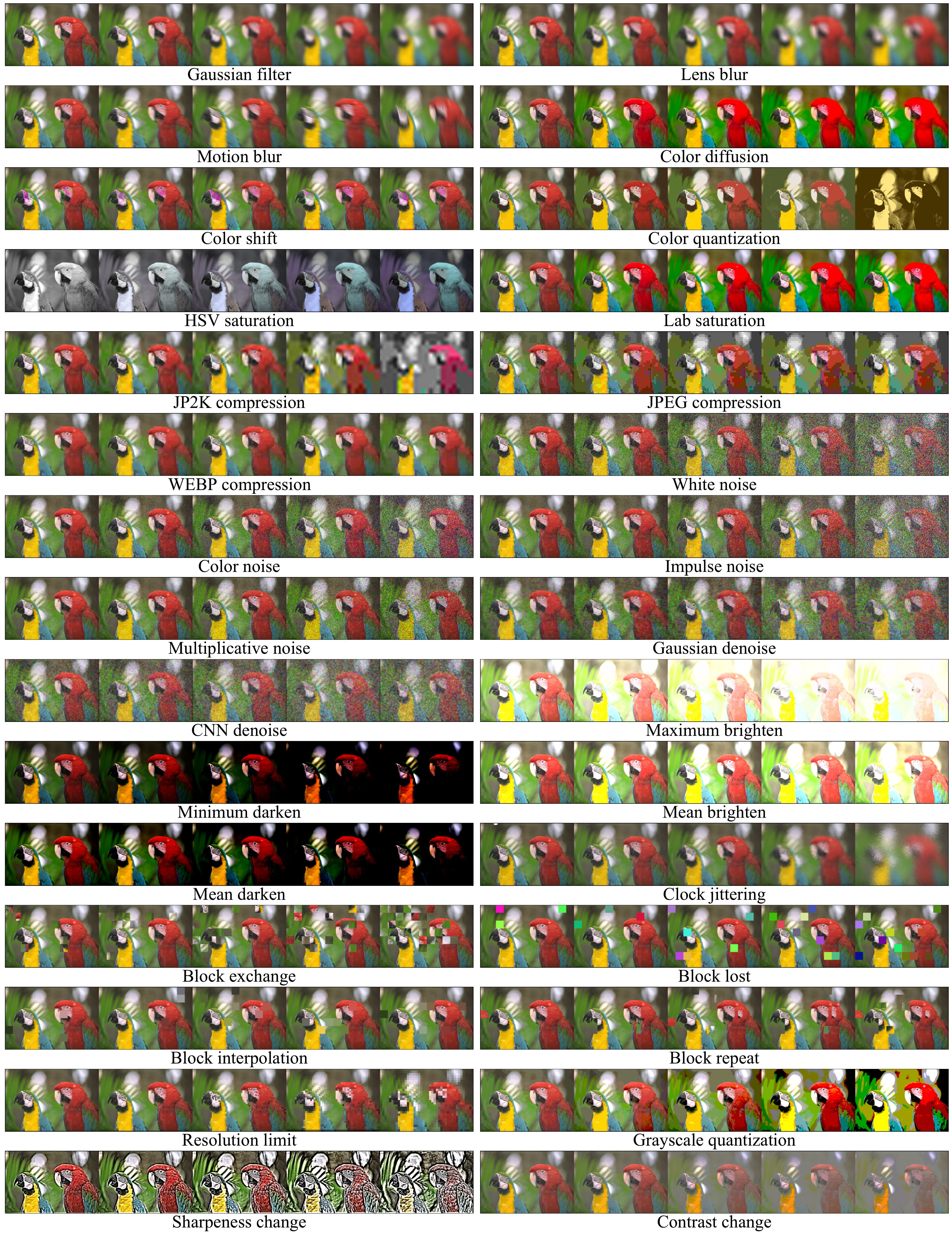}
\vspace{-7mm}
\caption{Visualization of 30 types of corruption, zoom in for detail. Strength from left (Level 1) to right (Level 5).}
\vspace{-7mm}
\label{fig:corruption}
\end{figure*}

Figure \ref{fig:corruption} shows 30 corruption types at different strength levels from 1 to 5. From a human’s subjective perspective, the visual quality of different corruptions is similar at the same strength. However, from this research, the perception of machines is completely inconsistent. This further proves the huge difference between MVS and HVS.

\section{Example od High\&Low Quality Instance}

Figures \ref{fig:low} and \ref{fig:high} show low-quality and high-quality examples, with the reference image on the left and the distorted image on the right. Overall, the machine is more robust to pixelation and block-level distortion but has a lower evaluation of grayscale and color changes. We hope that these data from MPD can further reveal the difference between HVS and MVS.

\section{Disclaimer}
The main purpose of this study is to propose the technical problem of IQA for Machine Vision, and it does not involve praising or criticizing any IQA models, nor is it a performance basis for IQA. Any IQA metric that performs poorly on MPD does not mean it is inferior. On the contrary, they all have excellent performance in past human-centric IQA while this study is the first machine-centric IQA task. The performance of these state-of-the-art models does not prove a poor capability, but the huge gap in preferences between humans and machines. We will release all data and annotations for non-commercial use, and we sincerely hope that future IQA developers will consider machine preference as an important issue to solve.

\section{Acknowledgment}

The work was supported by the National Natural Science Foundation of China under Grant 62225112, 62301310, 623B2073, 62471290; and in part by Sichuan Science and Technology Program under Grant 2024NSFSC1426.


{
    \small
    \bibliographystyle{ieeenat_fullname}
    \bibliography{main}
}

\end{document}